\documentclass[10pt,twocolumn,letterpaper]{article}

\usepackage{iccv}
\usepackage{times}
\usepackage{epsfig}
\usepackage{graphicx}
\usepackage{amsmath}
\usepackage{amssymb}
\usepackage{multirow}

\graphicspath{{./figures/}}

\usepackage{color}
\usepackage{cite}
\usepackage{dsfont}
\usepackage{amsmath}
\usepackage{amssymb}
\usepackage{amsfonts}
\usepackage{mathtools}
\usepackage{bm}
\usepackage[table,xcdraw]{xcolor}
\usepackage{ragged2e}
\usepackage{nccmath}
\usepackage{textcomp}
\usepackage{arydshln}
\usepackage{amsmath}
\usepackage{caption}
\usepackage{xfrac}

\newcommand{\mat}[1]{\bm{\mathrm{#1}}}
\newcommand{\light}{\ell}
\newcommand{\lossfun}{\mathcal{L}}
\newcommand{\headercolor}{FFFFFF}
\newcommand{\bestcolor}{FFFC9E}
\newcommand{\secondbestcolor}{FFCCC9}

\newcommand{\modelweights}{\theta}
\newcommand{\waugmentation}{w/aug.}
\newcommand{\numinputs}{m}
\newcommand{\sobel}{\nabla}

\usepackage[breaklinks=true,bookmarks=false]{hyperref}

\iccvfinalcopy 

\def\iccvPaperID{1277} 

\ificcvfinal\fi

\begin{document}

\title{Cross-Camera Convolutional Color Constancy}

\author{Mahmoud Afifi\textsuperscript{1,2}\thanks{This work was done while Mahmoud was an intern at Google.} \hspace{5mm} Jonathan T. Barron\textsuperscript{1} \hspace{5mm} Chloe LeGendre\textsuperscript{1} \hspace{5mm} Yun-Ta Tsai\textsuperscript{1} \hspace{5mm} Francois Bleibel\textsuperscript{1}
\vspace{5mm}\\
\textsuperscript{1}Google Research \hspace{10mm} \textsuperscript{2}York University}

\maketitle
\ificcvfinal\thispagestyle{empty}\fi

\begin{abstract}
We present ``Cross-Camera Convolutional Color Constancy'' (C5), a learning-based method, trained on images from multiple cameras, that accurately estimates a scene's illuminant color from raw images captured by a new camera previously unseen during training.
C5 is a hypernetwork-like extension of the convolutional color constancy (CCC) approach: C5 learns to generate the weights of a CCC model that is then evaluated on the input image, with the CCC weights dynamically adapted to different input content.
Unlike prior cross-camera color constancy models, which are usually designed to be agnostic to the spectral properties of test-set images from unobserved cameras, C5 approaches this problem through the lens of transductive inference: additional unlabeled images are provided as input to the model at test time, which allows the model to calibrate itself to the spectral properties of the test-set camera during inference.
C5 achieves state-of-the-art accuracy for cross-camera color constancy on several datasets, is fast to evaluate ($\sim$7 and $\sim$90 ms per image on a GPU or CPU, respectively), and requires little memory ($\sim$2 MB), and thus is a practical solution to the problem of calibration-free automatic white balance for mobile photography.
\end{abstract}

\section{Introduction} \label{sec.intro}

\begin{figure}
\centering
\includegraphics[width=\linewidth]{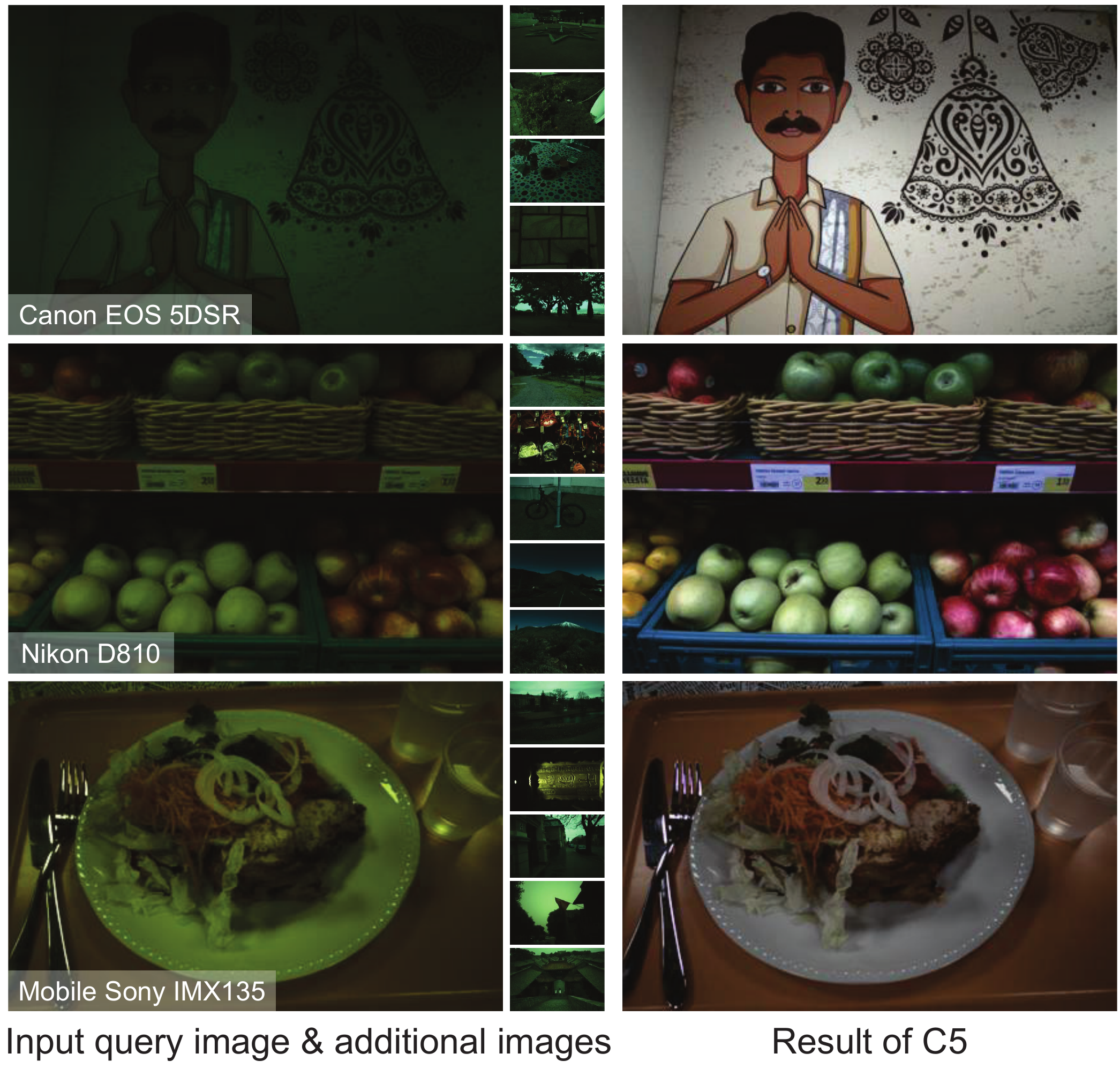}
\vspace{-15pt}
\caption{\small{Our C5 model exploits the colors of unlabeled additional images captured by the new camera model to generate a specific color constancy model for the input image.\ These additional images can be \textit{randomly} loaded from the photographer’s ``camera roll'', or they could be a fixed set taken once by the camera manufacturer. The shown images were captured by \textit{unseen} DSLR and smartphone camera models \cite{laakom2019intel} that were not included in the training stage. \label{fig:teaser}}}
\vspace{-10pt}
\end{figure}

The goal of computational color constancy is to emulate the human visual system's ability to constantly perceive object colors even when they are observed under different illumination conditions.\ In many contexts, this problem is equivalent to the practical problem of automatic white balance---removing an undesirable global color cast caused by the illumination in the scene, thereby, making it appear to have been imaged under a white light (see Figure\ \ref{fig:teaser}).\ White balance does not only affect the quality of photographs but also has an impact on the accuracy of different computer vision tasks  \cite{afifi2019else}.\ On modern digital cameras, automatic white balance is performed for all captured images as an essential part of the camera's imaging pipeline.

Color constancy is a challenging problem, because it is fundamentally under-constrained: an infinite family of white-balanced images and global color casts can explain the same observed image. Color constancy is, therefore, often framed in terms of inferring the most likely illuminant color given some observed image and some prior knowledge of the spectral properties of the camera's sensor.

One simple heuristic applied to the color constancy problem is the ``gray-world'' assumption: that colors in the world tend to be neutral gray and that the color of the illuminant can, therefore, be estimated as the average color of the input image~\cite{GW}.\ This gray-world method and its related techniques have the convenient property that they are \emph{invariant} to much of the spectral sensitivity differences among camera sensors and, therefore, very well-suited to the cross-camera task. If camera A's red channel is twice as sensitive as camera B's red channel, then a scene captured by camera A will have an average red intensity that is twice that of the scene captured by camera B, and so gray-world will produce identical output images (though this assumes that the spectral response of A and B are identical up to a scale factor, which is rarely the case in practice). However, current state-of-the-art learning-based methods for color constancy rarely exhibit this property, because they often learn things like the precise distribution of likely illuminant colors (a consequence of black-body illumination and other scene lighting regularities) and are, therefore, sensitive to any mismatch between the spectral sensitivity of the camera used during training and that of the camera used at test time \cite{afifi2019sensor}.

\begin{figure}
\centering
\includegraphics[width=0.8\linewidth]{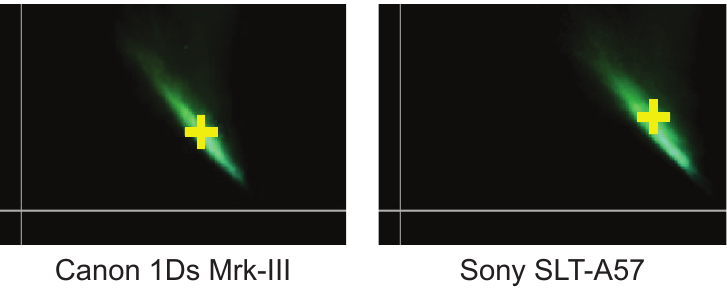}
\vspace{-5pt}
\caption{\small{
A visualization of $uv$ log-chroma histograms ($u = \log(g / r)$, $v = \log(g / b)$) of images from two different cameras averaged over many images of the same scene set in the NUS dataset~\cite{cheng2014illuminant} (shown in green), as well as the $uv$ coordinate of the mean of ground-truth illuminants over the entire scene set (shown in yellow).
The ``positions'' of these histograms change significantly across the two camera sensors because of their different spectral sensitivities, which is why many color constancy models generalize poorly across cameras.
\label{fig:idea}}}
\vspace{-10pt}
\end{figure}

Because there is often significant spectral variation across camera models (as shown in Figure~\ref{fig:idea}), this sensitivity of existing methods is problematic when designing practical white-balance solutions. Training a learning-based algorithm for a new camera requires collecting hundreds, or thousands, of images with ground-truth illuminant color labels (in practice: images containing a color chart), a burdensome task for a camera manufacturer or platform that may need to support hundreds of different camera models. However, the gray-world assumption still holds surprisingly well across sensors---if given several images from a particular camera, one can do a reasonable job of estimating the range of likely illuminant colors (as can also be seen in Figure~\ref{fig:idea}).

In this paper, we propose a camera-independent color constancy method. Our method achieves high-accuracy cross-camera color constancy through the use of two concepts: First, our system is constructed to take as input not just a single test-set image, but also a small set of additional images from the test set, which are: (i) arbitrarily-selected, (ii) unlabeled,  (iii) and not white balanced. This allows the model to calibrate itself to the spectral properties of the test-time camera during inference. We make \textit{no assumptions} about these additional images except that they come from the same camera as the ``target'' test set image and they contain some content (not all black or white images). In practice, these images could simply be \textit{randomly} chosen images from the photographer's ``camera roll'', or they could be a fixed set of ad hoc images of natural scenes taken once by the camera manufacturer---because these images do not need to be annotated, they are abundantly available. Second, our system is constructed as a \emph{hypernetwork}~\cite{ha2016hypernetworks} around an existing color constancy model. The target image and the additional images are used as input to a deep neural network whose output is the weights of a smaller color constancy model, and those generated weights are then used to estimate the illuminant color of the target image. 

Our system is trained using labeled (and unlabeled) images from multiple cameras, but at test time our model is able to look at a set of (unlabeled) test set images from a new camera. Our hypernetwork is able to infer the likely spectral properties of the new camera that produced the test set images (much as the reader can infer the likely illuminant colors of a camera from only looking at aggregate statistics, as in Figure~\ref{fig:idea}) and produce a small model that has been dynamically adapted to produce accurate illuminant estimates when applied to the target image. Our method is computationally fast and requires a low memory footprint while achieving state-of-the-art results compared to other camera-independent color constancy methods.

\section{Prior Work} \label{sec:prior}

There is a large body of literature proposed for illuminant color estimation, which can be categorized into statistical-based methods (e.g., \cite{GW, maxRGB, maxRGB_2, SoG, BrightPixels, GE, WGE, cheng2014illuminant, GI}) and learning-based methods (e.g., \cite{forsyth1990novel, PixelGamut, gamut, brainard1997bayesian, rosenberg2001color, gehler2008bayesian, bmvc1, bianco2015color, CCC, Seoung, shi2016deep, hu2017fc, FFCC, mcdonagh2018formulating, igtn}). The former rely on statistical-based hypotheses to estimate scene illuminant colors based on the color distribution and/or spatial layout of the input raw image. Such methods are usually simple and efficient, but they are less accurate than the learning-based alternatives. 

Learning-based methods, on the other hand, are typically trained for a single target camera model in order to learn the distribution of illuminant colors produced by the target camera's particular sensor \cite{afifi2019sensor, koskinen12cross, gao2017improving}. The learning-based methods are typically constrained to the specific, single camera use-case, as the spectral sensitivity of each camera sensor significantly alters the recorded illuminant and scene colors, and different sensor spectral sensitivities change the illuminant color distribution for the same set of scenes \cite{jiang2013space, wug2016yourself}. Such camera-specific methods cannot accurately extrapolate beyond the learned distribution of the training camera model's illuminant colors \cite{afifi2019sensor, GI} without tuning/re-training or pre-calibration \cite{liba2019handheld}.

Recently, few-shot and multi-domain learning techniques \cite{xiao2020multi, mcdonagh2018formulating} have been proposed to reduce the effort of re-training camera-specific learned color constancy models. These methods require only a small set of labeled images for a new camera unseen during training. In contrast, our technique requires no ground-truth labels for the unseen camera, and is essentially calibration-free for this new sensor. 

Another strategy has been proposed to white balance the input image with several illuminant color candidates and learn the likelihood of properly white-balanced images \cite{hernandez2020multi}. Such a Bayesian framework requires prior knowledge of the target camera model's illuminant colors to build the illuminant candidate set. Despite promising results, these methods, however, all require labeled training examples from the target camera model: raw images paired with ground-truth illuminant colors. Collecting such training examples is a tedious process, as certain conditions must be satisfied---i.e., for each image to have a single uniform lighting and a calibration object to be present in the scene \cite{cheng2014illuminant}.

An additional class of work has sought to learn \textit{sensor-independent} color constancy models, circumventing the need to re-train or calibrate to a specific camera model. A recent quasi-unsupervised approach to color constancy has been proposed, which learns the semantic features of achromatic objects to help build a model robust to differing camera sensor spectral sensitivities \cite{bianco2019quasi}. Another technique proposes to learn an intermediate ``device independent" space before the illuminant estimation process \cite{afifi2019sensor}. The goal of our method is similar, in that we also propose to learn a color constancy model that works for all cameras, but neither of these previous sensor-independent approaches leverages multiple test images to reason about the spectral properties of the unseen camera model. This enables our method to outperform these state-of-the-art sensor-independent methods across diverse test sets.

Though not commonly applied in color constancy techniques, our proposal to use multiple test-set images at inference-time to improve performance is a well-explored approach across machine learning. The task of classifying an entire test set as accurately as possible was first described by Vapnik as ``transductive inference'' \cite{vapnik1998statistical, transductive_1}.
Our approach is also closely related to the work on domain adaptation~\cite{daume2006domain, saenko2010adapting} and transfer learning~\cite{pan2009survey}, both of which attempt to enable learning-based models to cope with differences between training and test data.
Multiple sRGB camera-rendered images of the same scene have been used to estimate the response function of a given camera in the radiometric calibration literature \cite{grossberg2004modeling, kim2008radiometric}. In our method, however, we employ additional images to learn to extract informative cues about the spectral sensitivity of the camera capturing the input test image, without needing to capture the same scene multiple times.

\section{Method} \label{sec.method}

We call our system ``cross-camera convolutional color constancy'' (C5), because it builds upon the existing ``convolutional color constancy'' (CCC) model~\cite{CCC} and its successor ``fast Fourier color constancy'' (FFCC)~\cite{FFCC}, but embeds them in a multi-input hypernetwork to enable accurate cross-camera performance. These CCC/FFCC models work by learning to perform localization within a log-chroma histogram space, such as those shown in Figure~\ref{fig:idea}. 


Here, we present a convolutional color constancy model that is a simplification of those presented in the original work \cite{CCC} and its FFCC follow-up~\cite{FFCC}. This simple convolutional model will be a fundamental building block that we will use in our larger neural network. The image formation model behind CCC/FFCC (and most color constancy models) is that each pixel of the observed image is assumed to be the element-wise product of some ``true'' white-balanced image (or equivalently, the observed image if it were imaged under a white illuminant) and some illuminant color:
\begin{equation}
\forall_k\,\mat{c}^{(k)} = \mat{w}^{(k)} \circ \mat{\light}\,,
\end{equation}
where $\mat{c}^{(k)}$ is the observed color of pixel $k$, $\mat{w}^{(k)}$ is the true color of the pixel, and $\mat{\light}$ is the color of the illuminant, all of which are 3-vectors of RGB values.
Color constancy algorithms traditionally use the input image $\{\mat{c}^{(k)}\}$
to produce an estimate of the illuminant $\hat{\mat{\light}}$ that is then divided (element-wise) into each observed color to produce an estimate of the true color of each pixel, $\{ \hat{\mat{w}}^{(k)} \}$.

CCC defines two log-chroma measures for each pixel, which are simply the log of the ratio of two color channels:
\begin{equation}
u^{(k)} = \log\big( c^{(k)}_g / c^{(k)}_r\big), \quad v^{(k)} = \log\big(c^{(k)}_g / c^{(k)}_b\big)\,.
\end{equation}

As noted by Finlayson, this log-chrominance representation of color means that illuminant changes (i.e.\, element-wise scaling by $\mat\light$) can be modeled simply as additive offsets to this $uv$ representation~\cite{finlayson2001color}.
We then construct a 2D histogram of the log-chroma values of all pixels:
\begin{equation}
\resizebox{0.9\linewidth}{!}{$%
N_0(u,v) = \displaystyle \sum_k || \mat{c}^{(k)} ||_2 \left[ \left|u^{(k)} - u \right| \leq \epsilon \, \wedge\,  \left| v^{(k)} - v \right| \leq \epsilon \right]\,. \label{eq:hist}
$}
\end{equation}

This is simply a histogram over all $uv$ coordinates of size ($64 \times 64)$ written out using Iverson brackets, where $\epsilon$ is the width of a histogram bin, and where each pixel is weighted by its overall brightness under the assumption that bright pixels provide more actionable signal than dark pixels.
As was done in FFCC, we construct two histograms: one of pixel intensities, $N_0$, and one of gradient intensities, $N_1$. The latter is constructed analogously to Equation~\ref{eq:hist}.

These histograms of log-chroma values exhibit a useful property: element-wise multiplication of the RGB values of an image by a constant results in a \emph{translation} of the resulting log-chrominance histograms. The core insight of CCC is that this property allows color constancy to be framed as the problem of ``localizing'' a log-chroma histogram in this $uv$ histogram-space~\cite{CCC}---because every $uv$ location in $N$ corresponds to a (normalized) illuminant color, $\mat{\light}$, the problem of estimating $\mat{\light}$ is \emph{reducible} (in a computability sense) to the problem of estimating a $uv$ coordinate. This can be done by discriminatively training a ``sliding window'' classifier much as one might train, say, a face-detection system: the histogram is convolved with a (learned) filter and the location of the argmax is extracted from the filter response, and that argmax corresponds to $uv$ value that is (the inverse of) an estimated illumination location.

We adopt a simplification of the convolutional structure used by FFCC~\cite{FFCC}:
\begin{equation}
P = \operatorname{softmax}\bigg(B + \sum_i\big(N_i * F_i\big)\bigg), \label{eq:CCC}
\end{equation}
where $\{ F_i \}$ and $B$ are filters and a bias, respectively, which have the same shape as $N_i$. 
Each histogram, $N_i$, is convolved with each filter, $F_i$, and summed across channels (a ``conv'' layer). Then, the bias, $B$, is added to that summation, which collectively biases inference towards $uv$ coordinates that correspond to common illuminants, such as black body radiation.

As was done in FFCC, this convolution is accelerated through the use of FFTs, though, unlike FFCC, we use a non-wrapped histogram and thus non-wrapped filters and bias. 
This avoids the need for the complicated ``de-aliasing'' scheme used by FFCC which is not compatible with the convolutional neural network structure that we will later introduce.

The output of the softmax, $P$, is effectively a ``heat map'' of what illuminants are likely, given the distribution of pixel and gradient intensities reflected in $N$ and in the prior $B$, from which, we extract a ``soft argmax'' by taking the expectation of $u$ and $v$ with respect to $P$:
\begin{equation}
    \hat{\light}_u = \sum_{u,v} u P(u,v)\,,\quad
    \hat{\light}_v = \sum_{u,v} v P(u,v). \label{eq:argmax}
\end{equation}

Equation\ \ref{eq:argmax} is equivalent to estimating the mean of a fitted Gaussian, in the $uv$ space, weighted by $P$. Because the absolute scale of $\mat{\light}$ is assumed to be irrelevant or unrecoverable in the context of color constancy, after estimating $(\hat{\light}_u, \hat{\light}_v)$, we produce an RGB illuminant estimate, $\hat{\mat{\light}}$, that is simply the unit vector whose log-chroma values match
\mbox{our estimate}:
\begin{gather}
\hat{\mat{\light}} = \left( \exp\left(-\hat{\light}_u\right) / z, \, 1 / z, \, \exp\left(-\hat{\light}_v\right) / z \right), \\
z = \sqrt{\exp\left(-\hat{\light}_u\right)^2 + \exp\left(-\hat{\light}_v\right)^2 + 1}. \label{eq:light_rgb}
\end{gather}

A convolutional color constancy model is then trained by setting $\{ F_i \}$ and $B$ to be free parameters which are then optimized to minimize the difference between the predicted illuminant, $\hat{\mat{\light}}$, and the ground-truth illuminant, $\mat{\light}^*$.

\subsection{Architecture} \label{subsec:architecture}

\begin{figure}[t]
\centering
\includegraphics[width=\linewidth]{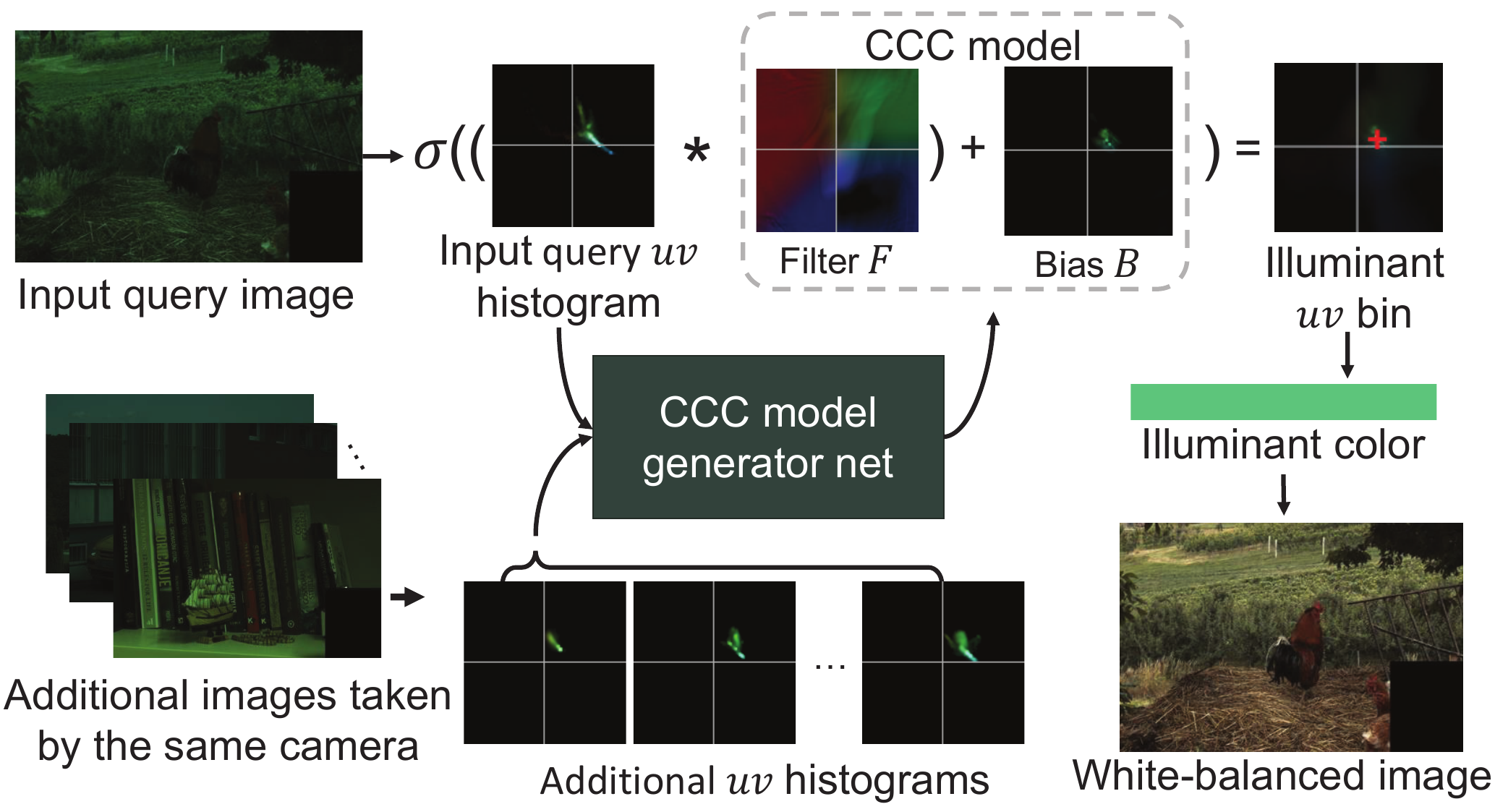}
\vspace{-15pt}
\caption{\small{
An overview of our C5 model.
The $uv$ histograms for the input query image and a variable number of additional input images taken from the same sensor as the query are used as input to our neural network, which generates a filter bank $\{F_i\}$ (here shown as one filter) and a bias $B$, which are the parameters of a conventional CCC model~\cite{CCC}.
The query $uv$ histogram is then convolved by the generated filter and shifted by the generated bias to produce a heat map, whose argmax is the estimated illuminant~\cite{CCC}.
\label{fig:main}}}
\vspace{-10pt}
\end{figure}

With our baseline CCC/FFCC-like model in place, we can now construct our cross-camera convolutional color constancy model (C5), which is a deep architecture in which CCC is a component. Both CCC and FFCC operate by learning a single fixed set of parameters consisting of a single filter bank $\{ F_i \}$ and bias $B$. In contrast, in C5 the filters and bias are parameterized as the output of a deep neural network (parameterized by weights $\modelweights$) that takes as input not just log-chrominance histograms for the image being color-corrected (which we will refer to as the ``query'' image), but also log-chrominance histograms from several other randomly selected input images (but with no ground-truth illuminant labels) from the test set.

By using a generated filter and bias from additional images taken from the query image's camera (instead of using a fixed filter and bias as was done in previous work) our model is able to automatically ``calibrate'' its CCC model to the specific sensor properties of the query image. This can be thought of as a hypernetwork~\cite{ha2016hypernetworks}, wherein a deep neural network emits the ``weights'' of a CCC model, which is itself a shallow neural network. This approach also bears some similarity to a Transformer approach, as a CCC model can be thought of as ``attending'' to certain parts of a log-chroma histogram, and so our neural network can be viewed as a sort of self-attention mechanism~\cite{vaswani2017attention}.
See Figure~\ref{fig:main} for a visualization of this data flow.

\begin{figure*}[ht]
\centering
\includegraphics[width=\linewidth]{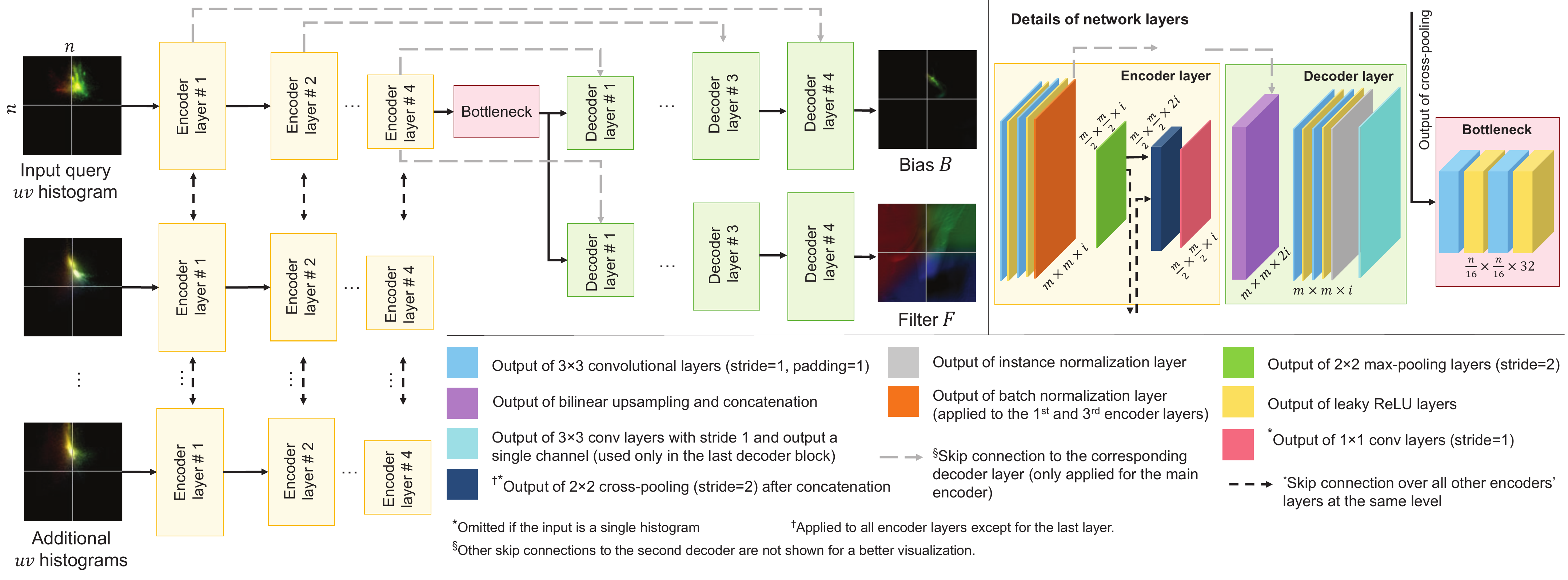}
\vspace{-15pt}
\caption{\small{An overview of neural network architecture that emits CCC model weights. The $uv$ histogram of the query image along with additional input histograms taken from the same camera are provided as input to a set of multiple encoders. The activations of each encoder are shared with the other encoders by performing max-pooling across encoders after each block. The cross-pooled features at the last encoder layer are then fed into two decoder blocks to generate a bias and filter bank of an CCC model for the query histogram. Each scale of the decoder is connected to the corresponding scale of the encoder for query histogram with skip connections. The structure of encoder and decoder blocks is shown at the upper right corner.\label{fig:architecture}}}
\vspace{-10pt}
\end{figure*}

At the core of our model is the deep neural network that takes as input a set of log-chroma histograms and must produce as output a CCC filter bank and bias map.  For this we use a multi-encoder-multi-decoder U-Net-like architecture \cite{ronneberger2015u}. The first encoder is dedicated to the ``query'' input image's histogram, while the rest of the encoders take as input the histograms corresponding to the additional input images. To allow the network to reason about the set of additional input images in a way that is insensitive to their ordering, we adopt the permutation invariant pooling approach of Aittala \etal \cite{aittala2018burst}: we use max pooling \emph{across} the set of activations of each branch of the encoder. This ``cross-pooling'' gives us a single set of activations that are reflective of the set of additional input images, but are agnostic to the particular ordering of those input images. At inference time, these additional images are needed to allow the network to reason about how to use them in challenging cases. 
The cross-pooled features of the last layer of all encoders are then fed into two decoder blocks. Each decoder produces one component of our CCC model: a bias map, $B$, and two filters, $\{ F_0, F_1 \}$ (which correspond to pixel and edge histograms, $\{ N_0, N_1 \}$, respectively).

As per the traditional U-Net structure, we use skip connections between each level of the decoder and its corresponding level of the encoder with the same spatial resolution, but only for the encoder branch corresponding to the query input image's histogram.
Each block of our encoder consists of a set of interleaved $3\times3$ conv layers, leaky ReLU activation, batch normalization, and $2\times2$ max pooling, and each block of our decoder consists of $2\times$ bilinear upsampling followed by interleaved $3\times3$ conv layers, leaky ReLU activation, and instance normalization.

When passing our 2-channel (pixel and gradient) log-chroma histograms to our network, we augment each histogram with two extra ``channels'' comprising of only the $u$ and $v$ coordinates of each histogram, as in CoordConv~\cite{liu2018intriguing}. This augmentation allows a convolutional architecture on top of log-chroma histograms to reason about the absolute ``spatial'' information associated with each $uv$ coordinate, thereby allowing a convolutional model to be aware of the absolute color of each histogram bin (see Appendix \ref{sec:ablation} for an ablation study).
Figure~\ref{fig:architecture} shows a detailed visualization of our architecture.

\subsection{Training} \label{subsec:training}

Our model is trained by minimizing the angular error \cite{hordley2004re} between the predicted unit-norm illuminant color, $\hat{\mat{\light}}$, and the ground-truth illuminant color, $\mat{\light}^*$, as well as an additional loss that regularizes the CCC models emitted by our network. Our loss function $\lossfun(\cdot)$ is:
\begin{equation}\label{loss:Eq.1}
\lossfun\left(\mat{\light}^*,  \hat{\mat{\light}} \right) = \cos^{-1}\left(\frac{\mat{\light}^* \cdot \hat{\mat{\light}}}{\lVert\mat{\light}^*\rVert}\right) + S\left(\{ F_i(\modelweights) \}, B(\modelweights)\right)\,,
\end{equation}
where $S(\cdot)$ is a regularizer that encourage the network to generate smooth filters and biases, which reduces over-fitting and improves generalization:
\begin{align}
S\left(\{ F_i \}, B\right) = \lambda_{B} (&\lVert B \ast \sobel_u \rVert^2 + \lVert B \ast \sobel_v \rVert^2 )  \nonumber \\
+ \lambda_{F} \sum_i (&\lVert F_i \ast \sobel_u \rVert^2 + \lVert F_i \ast \sobel_v \rVert^2 ) \,, \label{loss:Eq.2}
\end{align}
\noindent where $\sobel_u$ and $\sobel_v$ are $3\!\times\!3$ horizontal and vertical Sobel filters, respectively, and $\lambda_{F}$ and $\lambda_{B}$ are multipliers that control the strength of the smoothness for the filters and the bias, respectively. This regularization is similar to the total variation smoothness prior used by FFCC~\cite{FFCC}, though here we are imposing it on the filters and bias generated by a neural network, rather than on a single filter bank and bias map. We set the multiplier hyperparameters $\lambda_{F}$ and $\lambda_{B}$ to 0.15 and 0.02, respectively (see Appendix \ref{sec:ablation} for an ablation study).

In addition to regularizing the CCC model emitted by our network, we additionally regularize the weights of our network themselves, $\modelweights$, using L2 regularization (i.e., ``weight decay'') with a multiplier of $5\!\times\!10^{-4}$. This regularization of our network serves a different purpose than the regularization of the CCC models emitted by our network---regularizing $\{ F_i(\modelweights) \}$ and $B(\modelweights)$ prevents over-fitting by the CCC model emitted by our network, while regularizing $\modelweights$ prevents over-fitting by the model \emph{generating} those CCC models.

Training is performed using the Adam optimizer \cite{kingma2014adam}
with hyperparameters $\beta_1=0.9$, $\beta_2=0.999$, for 60 epochs. We use a learning rate of $5\!\times\!10^{-4}$ with a cosine annealing schedule \cite{loshchilov2016sgdr} and increasing batch-size (from 16 to 64) \cite{smith2017don, masters2018revisiting} which improve the stability of training (see Appendix \ref{sec:ablation} for an ablation study).

When training our model for a particular camera model, at each iteration we randomly select a batch of training images (and their corresponding ground-truth illuminants) for use as query input images, and then randomly select \textit{eight} additional input images for each query image from the training set for use as additional input images. See Appendix \ref{sec:additional_results} for results of multiple versions of our model in which we vary the number of additional images used.

\begin{figure}[]
\centering
\includegraphics[width=\linewidth]{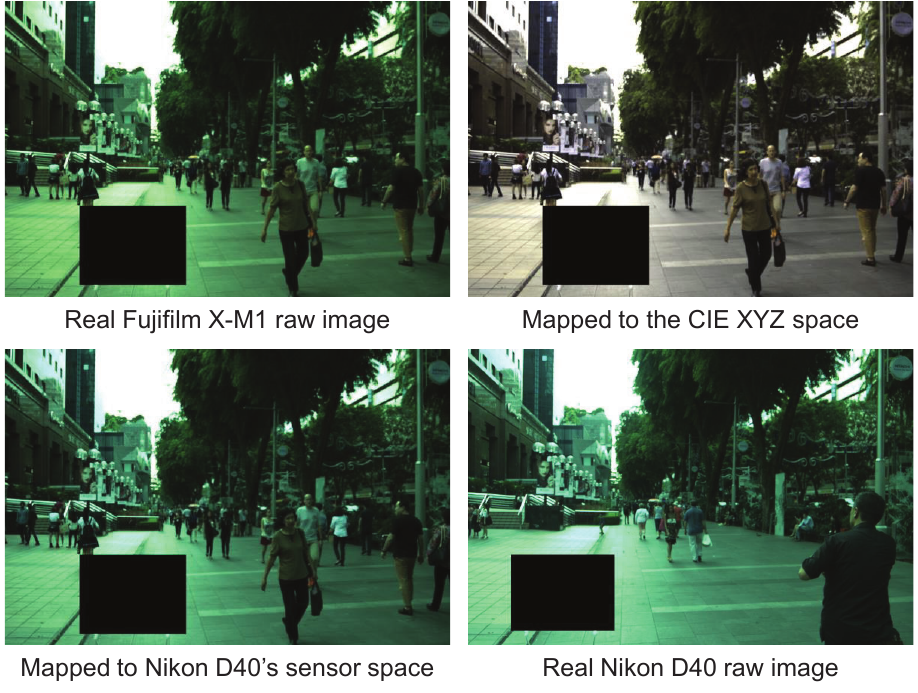}
\vspace{-15pt}
\caption{\small{An example of the image mapping used to augment training data. From left to right: a raw image captured by a Fujifilm X-M1 camera; the same image after white-balancing in CIE XYZ; the same image mapped into the Nikon D40 sensor space; and a real image captured by a Nikon D40 of the same scene for comparison \cite{cheng2014illuminant}.\label{fig:sensor_mapping_example}}}
\vspace{-10pt}
\end{figure}

\section{Experiments and Discussion} \label{sec:results}

In all experiments we used $384\!\times\!256$ raw images after applying the black-level normalization and masking out the calibration object to avoid any ``leakage'' during the evaluation. Excluding histogram computation time (which is difficult to profile accurately due to the expensive nature of scatter-type operations in deep learning frameworks), our method runs in $\sim$7 milliseconds per image on a NVIDIA GeForce GTX 1080, and $\sim$90 milliseconds on an Intel Xeon CPU Processor E5-1607 v4 (10M Cache, 3.10 GHz). Because our model exists in log-chroma histogram space, the uncompressed size of our entire model is $\sim$2 MB, small enough to easily fit within the narrow constraints of limited compute environments such as mobile phones.

\begin{figure*}[!t]
\centering
\includegraphics[width=\linewidth]{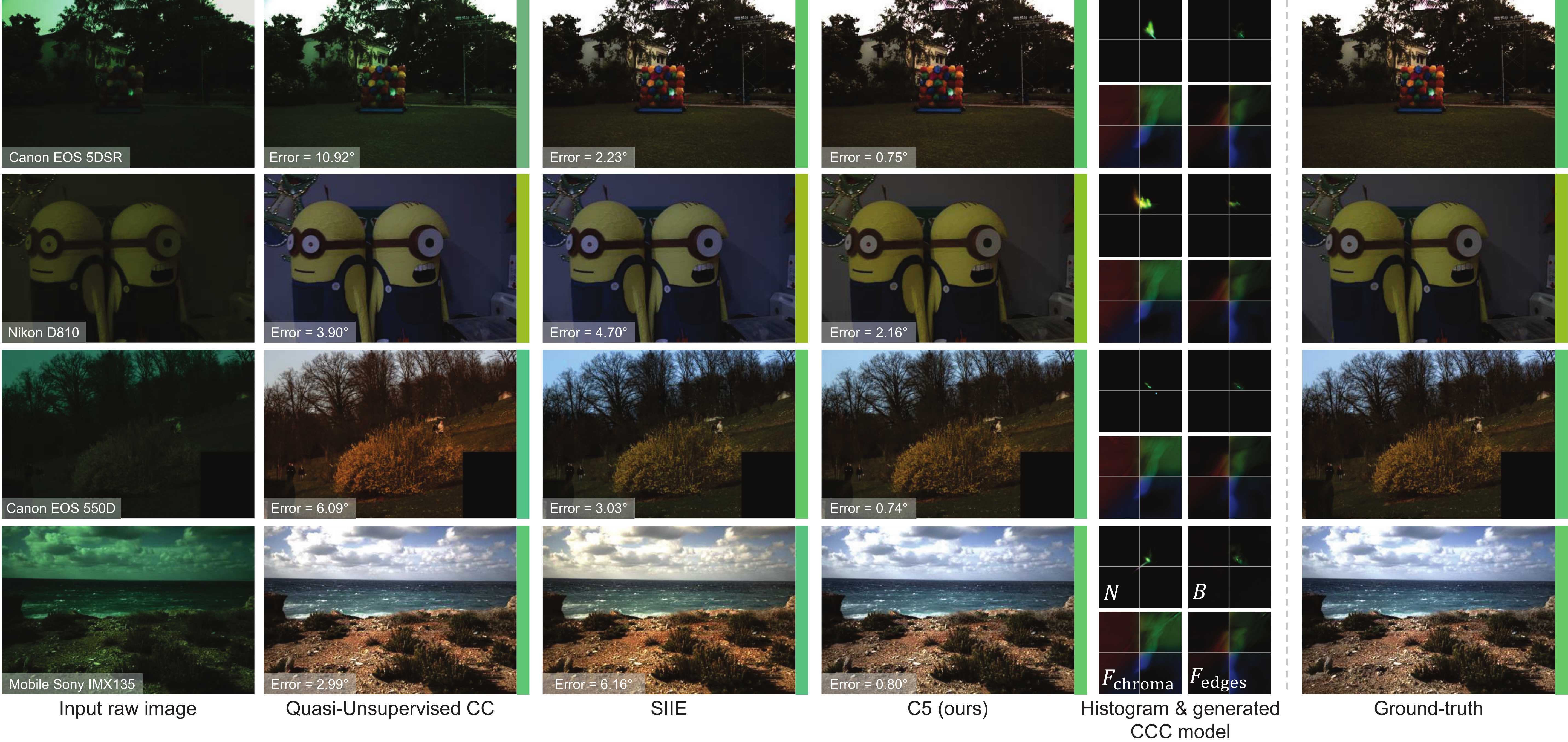}
\vspace{-15pt}
\caption{\small{
Here we visualize the performance of our C5 model alongside other camera-independent models: ``quasi-unsupervised CC'' \cite{bianco2019quasi} and SIIE \cite{afifi2019sensor}.
Despite not having seen any images from the test-set camera during training, C5 is able to produce accurate illuminant estimates. The intermediate CCC filters and biases produced by C5 are also visualized.\label{fig:qualitative_comparisons}}}
\vspace{-5pt}
\end{figure*}

\subsection{Data Augmentation}\label{subsec:dataaug}
Many of the datasets we use contain only a few images per distinct camera model (e.g. the NUS dataset \cite{cheng2014illuminant}) and this poses a problem for our approach as neural networks generally require significant amounts of training data. To address this, we use a data augmentation procedure in which images taken from a ``source'' camera model are mapped into the color space of a ``target'' camera.

To perform this mapping, we first white balance each raw source image using its ground-truth illuminant color, and then transform that white-balanced raw image into the device-independent CIE XYZ color space \cite{cie1932commission} using the color space transformation matrix (CST) provided in each DNG file \cite{DNG}. Then, we transform the CIE XYZ image into the target sensor space by inverting the CST of an image taken from the target camera dataset. 

Instead of randomly selecting an image from the target dataset, we use the correlated color temperature of each image and the capture exposure setting to match source and target images that were captured under roughly the same conditions. This means that ``daytime'' source images get warped into the color space of ``daytime'' target images, etc., and this significantly increases the realism of our synthesized data. After mapping the source image to the target white-balanced sensor space, we randomly sample from a cubic curve that has been fit to the $rg$ chromaticity of illuminant colors in the target sensor. 

Lastly, we apply a chromatic adaptation to generate the augmented image in the target sensor space. This chromatic adaptation is performed by multiplying each color channel of the white-balanced raw image, mapped to the target sensor space, with the corresponding sampled illuminant color channel value; see Figure \ref{fig:sensor_mapping_example} for an example.\ Additional details can be found in Appendix \ref{sec:augmentation}.\ This augmentation allows us to generate additional training examples to improve the generalization of our model. More details are provided in Sec.\ \ref{sbusec:results}.

\subsection{Results and Comparisons}\label{sbusec:results}

We validate our model using four public datasets consisting of images taken from one or more camera models: the Gehler-Shi dataset (568 images, two cameras) \cite{gehler2008bayesian}, the NUS dataset (1,736 images, eight cameras) \cite{cheng2014illuminant}, the INTEL-TAU dataset (7,022 images, three cameras) \cite{laakom2019intel}, and the Cube+ dataset (2,070 images, one camera) \cite{banic2017unsupervised} which has a separate 2019 ``Challenge'' test set~\cite{challenge}.
We measure performance by reporting the error statistics commonly used by the community: the mean, median, trimean, and arithmetic means of the first and third quartiles (``best 25\%'' and ``worst 25\%'') of the angular error between the estimated illuminant and the true illuminant. As our method randomly selects the additional images, each experiment is repeated ten times and we reported the arithmetic mean of each error metric (Appendix \ref{sec:additional_results} contains standard deviations).

\begin{figure}[]
\centering
\includegraphics[width=\linewidth]{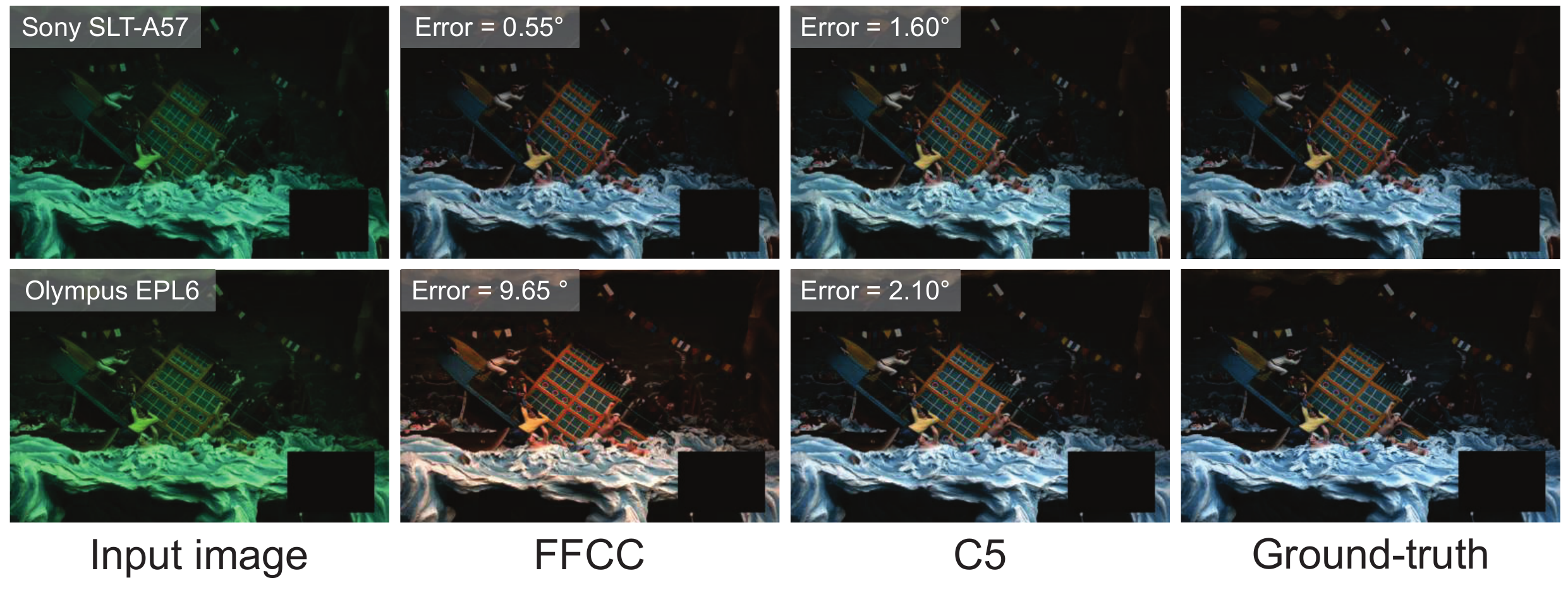}
\vspace{-11pt}
\small{
\caption{Here we compare our C5 model against FFCC~\cite{FFCC} on cross-sensor generalization using test-set Sony SLT-A57 images from the NUS dataset~\cite{cheng2014illuminant}. If FFCC is trained and tested on images from the same camera it performs well, as does C5 (top row). But if FFCC is instead tested on a different camera, such as the Olympus EPL6, it generalizes poorly, while C5 retains its performance (bottom row).\label{fig:comparison_w_FFCCC}}}
\vspace{-5pt}
\end{figure}

To evaluate our model's performance at generalizing to new camera models not seen during training, we adopt a leave-one-out cross-validation evaluation approach: for each dataset, we exclude all scenes and cameras used by the test set from our training images. For a fair comparison with FFCC \cite{FFCC}, we trained FFCC using the same leave-one-out cross-validation evaluation approach. Results can be seen in Table~\ref{table:results} and qualitative comparisons are shown in Figures~\ref{fig:qualitative_comparisons} and~\ref{fig:comparison_w_FFCCC}. Even when compared with prior sensor-independent techniques \cite{bianco2019quasi, afifi2019sensor}, we achieve state-of-the-art performance, as demonstrated in Table~\ref{table:results}.

When evaluating on the two Cube+ ~\cite{banic2017unsupervised,challenge} test sets and the INTEL-TAU~\cite{laakom2019intel} dataset in Table~\ref{table:results}, we train our model on the NUS \cite{cheng2014illuminant} and Gehler-Shi \cite{gehler2008bayesian} datasets. When evaluating on the Gehler-Shi \cite{gehler2008bayesian} and the NUS \cite{cheng2014illuminant} datasets in Table~\ref{table:results}, we train C5 using the INTEL-TAU dataset \cite{laakom2019intel}, the Cube+ dataset \cite{banic2017unsupervised}, and one of the Gehler-Shi \cite{gehler2008bayesian} and the NUS \cite{cheng2014illuminant} datasets after excluding the testing dataset. The one deviation from this procedure is for the NUS result labeled ``CS'', where for a fair comparison with the recent SIIE method \cite{afifi2019sensor}, we report our results with their cross-sensor (CS) evaluation in Table~\ref{table:results}, in which we only excluded images of the test camera, and repeated this process over all cameras in the dataset.

We augmented the data used to train the model, adding 5,000 augmented examples generated as described in Sec.\ \ref{subsec:dataaug}. In this process, we used only cameras of the training sets of each experiment as ``target" cameras for augmentation, which has the effect of mixing the sensors and scene content from the training sets only. For instance, when evaluating on the INTEL-TAU~\cite{laakom2019intel} dataset, our augmented images simulate the scene content of the NUS \cite{cheng2014illuminant} dataset as observed by sensors of the Gehler-Shi \cite{gehler2008bayesian} dataset, and vice-versa.

\paragraph{Characteristics of Additional Images}
Unless otherwise stated, the additional input images are randomly selected, but from the same camera model as the test image. This setting is meant to be equivalent to the real-world use case in which the additional images provided as input are, say, a photographer's previously-captured images that are already present on the camera during inference. However, for the ``Cube+ Challenge'' table, we provide an additional set of experiments in Table~\ref{table:results}, in which the set of additional images are chosen according to some heuristic, rather than randomly. We identified the 20 test-set images with the lowest variation of $uv$ chroma values (``dull images''), the 20 test-set images with the highest variation of $uv$ chroma values (``vivid images''), and we show that using vivid images produces lower error rates than randomly-chosen or dull images. This makes intuitive sense, as one might expect colorful images to be a more informative signal as to the spectral properties of previously-unobserved camera. We also show results in Table~\ref{table:results} where the additional images are taken from a different camera than the test-set camera, and show that this results in error rates that are higher than using additional images from the same test-set camera, as one might expect. 

\begin{table}[t]
\centering
\caption{Angular errors on the Cube+ dataset \cite{banic2017unsupervised}, the Cube+ challenge \cite{challenge}, the INTEL-TAU dataset \cite{laakom2019intel},  the Gehler-Shi dataset \cite{gehler2008bayesian}, and the NUS dataset \cite{cheng2014illuminant}. The term ``CS'' refers to cross-sensor as used in \cite{afifi2019sensor}. See the text for additional details. Lowest errors are highlighted in yellow.
\label{table:results}}
\vspace{0.5mm}
\resizebox{\linewidth}{!}{
\begin{tabular}{l|ccccc|c}
\textbf{Cube+ Dataset} & \textbf{Mean} & \textbf{Med.} & \textbf{B. 25\%} & \textbf{W. 25\%} & \textbf{Tri.} & \textbf{Size (MB)} \\ \hline

Gray-world \cite{GW}& 3.52 & 2.55 & 0.60 &  7.98 & 2.82 & - \\
Shades-of-Gray \cite{SoG}& 3.22 & 2.12 &  \cellcolor[HTML]{\bestcolor} 0.43 & 7.77 & 2.44 & - \\
Cross-dataset CC \cite{koskinen12cross} & 2.47 & 1.94  & - & - & - & - \\
Quasi-Unsupervised CC \cite{bianco2019quasi} & 2.69 & 1.76 & 0.49 & 6.45 & 2.00 & 622\\
SIIE \cite{afifi2019sensor} & 2.14 & 1.44 & 0.44 &  5.06 &  -
& 10.3 \\
FFCC \cite{FFCC} & 2.69 & 1.89 & 0.46 & 6.31 & 2.08 & 0.22 \\ \hdashline
C5 & \cellcolor[HTML]{\bestcolor} 1.92 & \cellcolor[HTML]{\bestcolor} 1.32 &  0.44 & \cellcolor[HTML]{\bestcolor} 4.44 & \cellcolor[HTML]{\bestcolor} 1.46 & 2.09 \\ 

\end{tabular}}

\vspace{0.11in}


\resizebox{\linewidth}{!}{
\begin{tabular}{l|ccccc}
\textbf{Cube+ Challenge} & \textbf{Mean} & \textbf{Med.} & \textbf{B. 25\%} & \textbf{W. 25\%} & \textbf{Tri.}  \\ \hline

Gray-world \cite{GW} & 4.44 & 3.50 & 0.77 & 9.64 & - \\
1st-order Gray-Edge \cite{GE} & 3.51 & 2.30 & 0.56 & 8.53 & - \\
Quasi-Unsupervised CC \cite{bianco2019quasi} & 3.12 & 2.19 & 0.60 & 7.28 & 2.40\\
SIIE \cite{afifi2019sensor} & 2.89 & 1.72 & 0.71 &  7.06 & - \\
FFCC \cite{FFCC} & 3.25 & 2.04 & 0.64 & 8.22 & 2.09 \\ \hdashline
C5 & 2.24 & 1.48 & 0.47 & \cellcolor[HTML]{\bestcolor}5.39 & 1.62 \\
C5 (another camera model) & 2.97 & 2.47 & 0.78 & 6.11 & 2.52\\
C5 (dull images) & 2.35 & 1.58 & 0.46 & 5.57 & 1.70 \\
C5 (vivid images) & \cellcolor[HTML]{\bestcolor}2.19 & \cellcolor[HTML]{\bestcolor}1.39 & \cellcolor[HTML]{\bestcolor}0.43 & 5.44 & \cellcolor[HTML]{\bestcolor}1.54 
\end{tabular}}

\vspace{0.11in}


\resizebox{\linewidth}{!}{
\begin{tabular}{l|ccccc}
\textbf{INTEL-TAU} & \textbf{Mean} & \textbf{Med.} & \textbf{B. 25\%} & \textbf{W. 25\%} & \textbf{Tri.}  \\ \hline

Gray-world \cite{GW} &  4.7 &  3.7 & 0.9 & 10.0 &  4.0\\
Shades-of-Gray \cite{SoG} & 4.0 &  2.9 & 0.7 & 9.0 &  3.2\\
PCA-based B/W Colors \cite{cheng2014illuminant} & 4.6 & 3.4 & 0.7 & 10.3 & 3.7 \\
Weighted Gray-Edge \cite{WGE} & 6.0 & 4.2 & 0.9 & 14.2 & 4.8\\
Quasi-Unsupervised CC \cite{bianco2019quasi} & 3.71 & 2.67 & 0.66 & 8.55 & 2.90\\
SIIE \cite{afifi2019sensor} & 3.42 & 2.42 & 0.73 &  7.80 & 2.64 \\
FFCC \cite{FFCC} & 3.42 & 2.38 & 0.70 & 7.96 & 2.61 \\ \hdashline
C5 & \cellcolor[HTML]{\bestcolor}2.52 & \cellcolor[HTML]{\bestcolor}1.70 & \cellcolor[HTML]{\bestcolor}0.52 & \cellcolor[HTML]{\bestcolor}5.96 & \cellcolor[HTML]{\bestcolor}1.86 \\
\end{tabular}}

\vspace{0.11in}


\resizebox{\linewidth}{!}{
\begin{tabular}{l|ccccc}
\textbf{Gehler-Shi Dataset} & \textbf{Mean} & \textbf{Med.} & \textbf{B. 25\%} & \textbf{W. 25\%} & \textbf{Tri.}  \\ \hline

Shades-of-Gray \cite{SoG}& 4.93 & 4.01 & 1.14 & 10.20  &  4.23 \\
PCA-based B/W Colors \cite{cheng2014illuminant}& 3.52 & 2.14 & 0.50 & 8.74  & 2.47 \\
ASM \cite{akbarinia2017colour} & 3.80 & 2.40 & - & - & 2.70 \\
Woo \textit{et al.} \cite{8226796} & 4.30 & 2.86 & 0.71 & 10.14 & 3.31 \\
Grayness Index \cite{GI} & 3.07 & 1.87 &  \cellcolor[HTML]{\bestcolor}0.43 &  7.62 &  2.16 \\
Cross-dataset CC \cite{koskinen12cross} & 2.87 & 2.21 & - & - & - \\
Quasi-Unsupervised CC \cite{bianco2019quasi} & 3.46 & 2.23 & - & - & - \\
SIIE \cite{afifi2019sensor} & 2.77 & \cellcolor[HTML]{\bestcolor}1.93 & 0.55 & 6.53 &  -  \\
FFCC \cite{FFCC} & 2.95 & 2.19 & 0.57 & 6.75 & 2.35 \\\hdashline 
CS & \cellcolor[HTML]{\bestcolor}2.50 & 1.99 & 0.53 & \cellcolor[HTML]{\bestcolor}5.46 & \cellcolor[HTML]{\bestcolor}2.03 \\ 
\end{tabular}
}

\vspace{0.11in}


\resizebox{\linewidth}{!}{
\begin{tabular}{l|ccccc}
\textbf{NUS Dataset} & \textbf{Mean} & \textbf{Med.} & \textbf{B. 25\%} & \textbf{W. 25\%} & \textbf{Tri.}  \\ \hline

Gray-world \cite{GW} & 4.59 & 3.46 & 1.16 & 9.85  & 3.81 \\
Shades-of-Gray \cite{SoG}& 3.67 & 2.94 & 0.98 &  7.75  & 3.03 \\
Local Surface Reflectance \cite{gao2014efficient}& 3.45 & 2.51 & 0.98 & 7.32  & 2.70 \\
PCA-based B/W Colors \cite{cheng2014illuminant}& 2.93 & 2.33 & 0.78 & 6.13 & 2.42 \\
Grayness Index \cite{GI} & 2.91 & 1.97 &  0.56 & 6.67 &  2.13 \\
Cross-dataset CC \cite{koskinen12cross} & 3.08 & 2.24 & - & - & - \\
Quasi-Unsupervised CC \cite{bianco2019quasi} & 3.00 & 2.25  & - & - & - \\	
SIIE (CS) \cite{afifi2019sensor} & 2.05 & 1.50 & 0.52 & 4.48  & - \\
FFCC \cite{FFCC} & 2.87 & 2.14 & 0.71 & 6.23 & 2.30
\\ \hdashline
CS & 2.54 & 1.90 & 0.61 & 5.61 & 2.02 \\ 
C5 (CS) & \cellcolor[HTML]{\bestcolor}1.77 & \cellcolor[HTML]{\bestcolor}1.37 & \cellcolor[HTML]{\bestcolor}0.48 & \cellcolor[HTML]{\bestcolor}3.75 & \cellcolor[HTML]{\bestcolor}1.46 \\
\end{tabular}
}\vspace{-4mm}
\end{table}

\section{Conclusion} \label{sec.conclusion}
We have presented C5, a cross-camera convolutional color constancy method. By embedding the existing state-of-the-art convolutional color constancy model (CCC) \cite{CCC, FFCC} into a multi-input hypernetwork approach, C5 can be trained on images from multiple cameras, but at test time synthesize weights for a CCC-like model that is dynamically calibrated to the spectral properties of the previously-unseen camera of the test-set image.
Extensive experimentation demonstrates that C5 achieves state-of-the-art performance on cross-camera color constancy for several datasets.
By enabling accurate illuminant estimation without requiring the tedious collection of labeled training data for every particular camera, we hope that C5 will accelerate the widespread adoption of learning-based white balance by the camera industry.

\appendix

\section{CCC Histogram Features}\label{appendix:histogram}

As mentioned earlier, we used a histogram bin size of 64 (i.e., $n=64$) with a histogram bin width $\epsilon = (b_{\texttt{max}} - b_{\texttt{min}})/n$, where $b_{\texttt{max}}$ and $b_{\texttt{min}}$ are the histogram boundary values. In our experiments, we set $b_{\texttt{min}}$ and $b_{\texttt{max}}$ to -2.85 and 2.85, respectively.\ Our input is a concatenation of two histograms: (i) a histogram of pixel intensities and (ii) a histogram of gradient intensities.\ We augmented our histograms with extra $uv$ coordinate channels to allow our network to consider the ``spatial'' (or more accurately, chromatic) information associated with each bin in the histogram.

\section{Ablations Studies}\label{sec:ablation}

In the following ablation experiments, we used the Cube+ dataset \cite{banic2017unsupervised} as our test set and trained our network with seven encoders using the same training set mentioned in Sec.\ \ref{sbusec:results} (the NUS dataset \cite{cheng2014illuminant}, the Gehler-Shi dataset \cite{gehler2008bayesian}, and the augmented images after excluding any scene/sensors of the test set). Table \ref{table:ablation} shows the results obtained by models trained using different histogram sizes, using different values of the smoothness factors $\lambda_{B}$ and $\lambda_{F}$, with and without increasing the batch-size during training, and with and without the histogram gradient intensity and the extra $uv$ augmentation channels. Each experiment was repeated ten times and the arithmetic mean and standard deviation of each error metric are reported.

\begin{figure*}
\centering
\includegraphics[width=\linewidth]{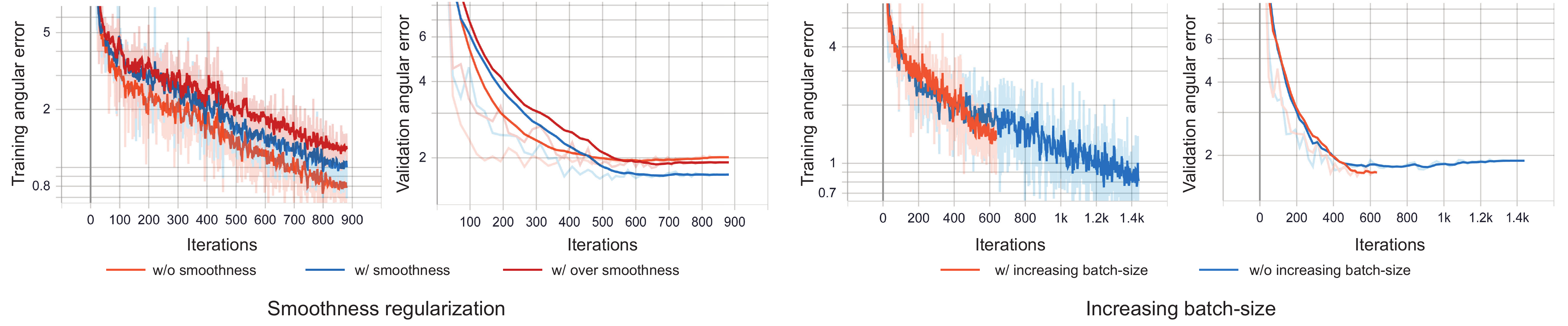}
\vspace{-6mm}
\caption{The impact of  smoothness regularization and of increasing the batch size during training on training/validation accuracy. We show the training/validation angular error of training our network on the Gehler-Shi dataset \cite{gehler2008bayesian} for camera-specific color constancy. We set $\lambda_{F}=0.15$, $\lambda_{B}=0.02$ for the experiment labeled with `w/ smoothness', while we used $\lambda_{F}=1.85$, $\lambda_{B}=0.25$ for the experiment labeled with `over smoothness' and $\lambda_{F}=0$, $\lambda_{B}=0$ for the `w/o smoothness' experiments.}
\label{fig:regularization_ablation}
\end{figure*}

Figure \ref{fig:regularization_ablation} shows the effect of the smoothness regularization and increasing the batch-size during training on a small training set. We use the first fold of the Gehler-Shi dataset \cite{gehler2008bayesian} as our validation set and the remaining two folds are used for training. In the figure, we plot the angular error on the training and validation sets. Each model was trained for 60 epochs as a camera-specific color constancy model (i.e., without using additional images or camera models). As can be seen in Figure \ref{fig:regularization_ablation}, the smoothness regularization improves the generalization on the test set and increasing the batch size helps the network to reach a lower optimum.

\begin{table*}[]

\centering
\caption{Results of ablation studies. The shown results were obtained by training our network on the NUS \cite{cheng2014illuminant} and the Gehler-Shi datasets \cite{gehler2008bayesian} with augmentation, and testing on the Cube+ dataset \cite{banic2017unsupervised}. In this set of experiments, we used seven encoders (i.e., six additional histograms). Note that none of the training data includes any scene/sensor from the Cube+ dataset \cite{banic2017unsupervised}. For each set of experiments, we highlight the lowest errors in yellow.\label{table:ablation}}
\vspace{-2mm}
\scalebox{0.75}{

\begin{tabular}{lccccc}
& \textbf{Mean} & \textbf{Med.} & \textbf{B. 25\%} & \textbf{W. 25\%} & \textbf{Tri.} \\ \cline{2-6} 
& \multicolumn{5}{c}{\cellcolor[HTML]{\headercolor}Histogram bin size, $n$} \\ \hline
\multicolumn{1}{l|}{$n=16$} &  2.28$\pm$0.01 & 1.81$\pm$0.03 & 0.65$\pm$0.01 & 4.72$\pm$0.02 & 1.91$\pm$0.02\\ \hline
\multicolumn{1}{l|}{$n=32$} &  2.02$\pm$0.01 & 1.44$\pm$0.01 & 0.44$\pm$0.01 & 4.66$\pm$0.01 & 1.86$\pm$0.03  \\ \hline
\multicolumn{1}{l|}{$n=64$} &  \cellcolor[HTML]{\bestcolor}1.87$\pm$0.00 & \cellcolor[HTML]{\bestcolor}1.27$\pm$0.01 & 0.41$\pm$0.01 & \cellcolor[HTML]{\bestcolor}4.36$\pm$0.01 & \cellcolor[HTML]{\bestcolor}1.40$\pm$0.01\\ \hline
\multicolumn{1}{l|}{$n=128$} & 2.03$\pm$0.00 & 1.42$\pm$0.01 & \cellcolor[HTML]{\bestcolor}0.40$\pm$0.00 & 4.70$\pm$0.01 & 1.54$\pm$0.01 \\ \hline

& \multicolumn{5}{c}{\cellcolor[HTML]{\headercolor}Smoothness factors, $\lambda_{B}$ and $\lambda_{F}$ ($n=64$)} \\ \hline
\multicolumn{1}{l|}{$\lambda_B=0, \lambda_F = 0$} & 2.07$\pm$0.01 & 1.42$\pm$0.01 & 0.47$\pm$0.01 & 4.67$\pm$0.01 & 1.57$\pm$0.01  \\ \hline

\multicolumn{1}{l|}{$\lambda_B=0.005, \lambda_F = 0.035$} & 1.95$\pm$0.00 & 1.31$\pm$0.01 & \cellcolor[HTML]{\bestcolor}0.40$\pm$0.00 & 4.57$\pm$0.01 & 1.47$\pm$0.01\\ \hline

\multicolumn{1}{l|}{$\lambda_B=0.02, \lambda_F = 0.15$} &  \cellcolor[HTML]{\bestcolor}1.87$\pm$0.00 & \cellcolor[HTML]{\bestcolor}1.27$\pm$0.01 & 0.41$\pm$0.01 & \cellcolor[HTML]{\bestcolor}4.36$\pm$0.01 & \cellcolor[HTML]{\bestcolor}1.40$\pm$0.01 \\ \hline
\multicolumn{1}{l|}{$\lambda_B=0.10, \lambda_F = 0.75$} & 2.11$\pm$0.00 & 1.55$\pm$0.01 & 0.48$\pm$0.00 & 4.70$\pm$0.01 & 1.66$\pm$0.01 \\ \hline

\multicolumn{1}{l|}{$\lambda_B=0.25, \lambda_F = 1.85$} & 2.23$\pm$0.00 & 1.61$\pm$0.01 & 0.53$\pm$0.00 & 5.04$\pm$0.01 & 1.77$\pm$ 0.01  \\ \hline

& \multicolumn{5}{c}{\cellcolor[HTML]{\headercolor}Increasing batch size ($n=64$)} \\ \hline

\multicolumn{1}{l|}{w/o increasing} & 1.93$\pm$0.00 & 1.29$\pm$0.01 & 0.42$\pm$0.00 & 4.52$\pm$0.02 & 1.43$\pm$0.01 \\ \hline
\multicolumn{1}{l|}{w/ increasing} &  \cellcolor[HTML]{\bestcolor}1.87$\pm$0.00 & \cellcolor[HTML]{\bestcolor}1.27$\pm$0.01 & \cellcolor[HTML]{\bestcolor}0.41$\pm$0.01 & \cellcolor[HTML]{\bestcolor}4.36$\pm$0.01 &  \cellcolor[HTML]{\bestcolor} 1.40$\pm$0.01  \\ \hline

& \multicolumn{5}{c}{\cellcolor[HTML]{\headercolor}Gradient histogram and $uv$ channels ($n=64$)} \\ \hline

\multicolumn{1}{l|}{w/o gradient histogram} & 2.30$\pm$0.01 & 1.53$\pm$0.01 & 0.45$\pm$0.01 & 5.51$\pm$0.02 & 1.71$\pm$0.02 \\ \hline
\multicolumn{1}{l|}{w/o $uv$} & 2.03$\pm$0.01 & 1.45$\pm$0.01 & 0.44$\pm$0.01 & 4.63$\pm$0.02 & 1.56$\pm$0.01  \\ \hline
\multicolumn{1}{l|}{w/ $uv$ and gradient histogram} &  \cellcolor[HTML]{\bestcolor}1.87$\pm$0.00 & \cellcolor[HTML]{\bestcolor}1.27$\pm$0.01 & \cellcolor[HTML]{\bestcolor}0.41$\pm$0.01 & \cellcolor[HTML]{\bestcolor}4.36$\pm$0.01 & \cellcolor[HTML]{\bestcolor}1.40$\pm$0.01  \\

\end{tabular}}
\end{table*}

Table \ref{table:data_augmentation} shows the results with and without using our data augmentation approach. The experiments labeled ``w/aug" in Table~\ref{table:data_augmentation} refer to using our data augmentation approach, as described in Sec.\ \ref{subsec:dataaug}. Additional details on the data augmentation process are given in Sec.\ \ref{sec:augmentation}.

\begin{table}[]
\centering
\caption{Angular errors on the Cube+ dataset \cite{banic2017unsupervised} and the INTEL-TAU dataset \cite{laakom2019intel}. In this experiment, we used six additional images (i.e., $\numinputs=7$) for our C5. Lowest errors are highlighted in yellow.
\label{table:data_augmentation}}
\resizebox{\linewidth}{!}{
\begin{tabular}{l|ccccc}
\textbf{Cube+ Dataset} & \textbf{Mean} & \textbf{Med.} & \textbf{B. 25\%} & \textbf{W. 25\%} & \textbf{Tri.} \\ \hline
Cross-dataset CC \cite{koskinen12cross} & 2.47 & 1.94  & - & - & - \\
Quasi-Unsupervised CC \cite{bianco2019quasi} & 2.69 & 1.76 & 0.49 & 6.45 & 2.00 \\
SIIE \cite{afifi2019sensor} & 2.14 & 1.44 & 0.44 &  5.06 &  - \\
FFCC \cite{FFCC} & 2.69 & 1.89 & 0.46 & 6.31 & 2.08 \\ \hdashline
C5 & 2.10 & 1.38 & 0.49 & 4.97 & 1.56 \\
C5 (\waugmentation) & \cellcolor[HTML]{\bestcolor} 1.87 & \cellcolor[HTML]{\bestcolor} 1.27 & \cellcolor[HTML]{\bestcolor} 0.41 & \cellcolor[HTML]{\bestcolor} 4.36 & \cellcolor[HTML]{\bestcolor} 1.40 \\

\end{tabular}}

\vspace{0.11in}


\resizebox{\linewidth}{!}{
\begin{tabular}{l|ccccc}
\textbf{INTEL-TAU} & \textbf{Mean} & \textbf{Med.} & \textbf{B. 25\%} & \textbf{W. 25\%} & \textbf{Tri.}  \\ \hline
Quasi-Unsupervised CC \cite{bianco2019quasi} & 3.71 & 2.67 & 0.66 & 8.55 & 2.90 \\
SIIE \cite{afifi2019sensor} & 3.42 & 2.42 & 0.73 &  7.80 & 2.64 \\
FFCC \cite{FFCC} & 3.42 & 2.38 & 0.70 & 7.96 & 2.61 \\ \hdashline
C5  & 2.62 & 1.85 & 0.54 & 6.05 & 2.00\\
C5 (\waugmentation) & \cellcolor[HTML]{\bestcolor}2.49 & \cellcolor[HTML]{\bestcolor}1.66 & \cellcolor[HTML]{\bestcolor}0.51 & \cellcolor[HTML]{\bestcolor}5.93 & \cellcolor[HTML]{\bestcolor}1.83 \\

\end{tabular}}
\end{table}

\section{Additional Results}\label{sec:additional_results}

In Table \ref{table:results}, we reported our results using eight additional images. In Table \ref{table:additional_images}, we report multiple versions of our model in which we vary $\numinputs$, the number of input images (and encoders) used ($\numinputs=1$ means that only the query image is used as an input with no additional images). Note that the single-image results ($\numinputs=1$) are not intended to be the central contribution of this work---they are provided \textit{only} as a point of comparison.



\begin{table}[]
\centering
\caption{Results using different number of the additional images (i.e., different values of $m$). Note that $m=7$, for example, means that we use six additional images along with the input image. For each experiment, we used the same training data explained in the main paper with augmentation. Lowest errors are highlighted in yellow. 
\label{table:additional_images}.}

\resizebox{0.82\linewidth}{!}{
\begin{tabular}{l|cccc}
\textbf{Cube+ Dataset} & \textbf{Mean} & \textbf{Med.} & \textbf{B. 25\%} & \textbf{W. 25\%} \\ \hline

C5 ($\numinputs=1$) & 2.60 & 1.86 & 0.55 & 5.89  \\
C5 ($\numinputs=3$) & 2.28 & 1.50 & 0.59 & 5.19 \\
C5 ($\numinputs=5$) & 2.23 & 1.52 & 0.56 & 5.11  \\
C5 ($\numinputs=7$) & \cellcolor[HTML]{\bestcolor} 1.87 & \cellcolor[HTML]{\bestcolor} 1.27 &  0.41 & \cellcolor[HTML]{\bestcolor} 4.36 \\
C5 ($\numinputs=9$) & 1.92 & 1.32 & 0.44 & 4.44 \\ 
C5 ($\numinputs=11$) & 1.93 & 1.41 & 0.42 & 4.35 \\
C5 ($\numinputs=13$) & 1.95 & 1.35 & \cellcolor[HTML]{\bestcolor} 0.40 &  4.52 \\

\end{tabular}}

\vspace{0.11in}


\resizebox{0.82\linewidth}{!}{
\begin{tabular}{l|cccc}
\textbf{Cube+ Challenge} & \textbf{Mean} & \textbf{Med.} & \textbf{B. 25\%} & \textbf{W. 25\%}  \\ \hline

C5 ($\numinputs=1$) & 2.70 & 2.00 & 0.61  & 6.15 \\
C5 ($\numinputs=7$) & 2.55 & 1.63 & 0.54 & 6.21   \\
C5 ($\numinputs=9$) & \cellcolor[HTML]{\bestcolor} 2.24 & \cellcolor[HTML]{\bestcolor} 1.48 & \cellcolor[HTML]{\bestcolor} 0.47 & \cellcolor[HTML]{\bestcolor}5.39 \\
C5 ($\numinputs=11$) & 2.41 & 1.72 & 0.54 & 5.58 \\
C5 ($\numinputs=13$) & 2.39 & 1.61 & 0.53 &  5.64 \\
\end{tabular}}

\vspace{0.11in}

\resizebox{0.82\linewidth}{!}{
\begin{tabular}{l|cccc}
\textbf{INTEL-TAU} & \textbf{Mean} & \textbf{Med.} & \textbf{B. 25\%} & \textbf{W. 25\%} \\ \hline

C5 ($\numinputs=1$) & 2.99 & 2.18 & 0.66  & 6.71\\
C5 ($\numinputs=7$) & \cellcolor[HTML]{\bestcolor}2.49 & \cellcolor[HTML]{\bestcolor}1.66 & \cellcolor[HTML]{\bestcolor}0.51 & \cellcolor[HTML]{\bestcolor}5.93 \\
C5 ($\numinputs=9$) & 2.52 & 1.70 & 0.52 & 5.96 \\
C5 ($\numinputs=11$) & 2.60 & 1.79 &  0.54 &  6.07 \\
C5 ($\numinputs=13$) & 2.57 & 1.74 & 0.52 & 6.08 \\
\end{tabular}}

\vspace{0.11in}


\resizebox{0.82\linewidth}{!}{
\begin{tabular}{l|cccc}
\textbf{Gehler-Shi Dataset} & \textbf{Mean} & \textbf{Med.} & \textbf{B. 25\%} & \textbf{W. 25\%}  \\ \hline

C5 ($\numinputs=1$) & 2.98 & 2.05 & 0.54 & 7.13 \\
C5 ($\numinputs=7$) & \cellcolor[HTML]{\bestcolor}2.36 & \cellcolor[HTML]{\bestcolor}1.61 & \cellcolor[HTML]{\bestcolor}0.44 & 5.60 \\
CS ($\numinputs=9$) & 2.50 & 1.99 & 0.53 & \cellcolor[HTML]{\bestcolor}5.46  \\ 
C5 ($\numinputs=11$) & 2.55 & 1.88 & 0.50 & 5.77 \\
C5 ($\numinputs=13$) & 2.46 & 1.74 & 0.50 & 5.73 \\
\end{tabular}
}

\vspace{0.11in}

\resizebox{0.82\linewidth}{!}{
\begin{tabular}{l|ccccc}
\textbf{NUS Dataset} & \textbf{Mean} & \textbf{Med.} & \textbf{B. 25\%} & \textbf{W. 25\%}  \\ \hline

C5 ($\numinputs=1$) & 2.84 & 2.20 & 0.69 & 6.14 \\
C5 ($\numinputs=7$) & 2.68 & 2.00 & 0.66 & 5.90 \\
CS ($\numinputs=9$) & 2.54 & 1.90 & \cellcolor[HTML]{\bestcolor}0.61 & 5.61 \\ 
C5 ($\numinputs=11$) & 2.64 & 1.99 & 0.65 & 5.75\\
C5 ($\numinputs=13$) & \cellcolor[HTML]{\bestcolor}2.49  & \cellcolor[HTML]{\bestcolor}1.88 & \cellcolor[HTML]{\bestcolor}0.61 & \cellcolor[HTML]{\bestcolor}5.43 \\
\end{tabular}
}
\end{table}

We did not include the ``gain'' multiplier, originally proposed in FFCC \cite{FFCC}, in our method elaborated in Sec. \ref{sec.method}, as it did not result in a consistent improved performance over all error metrics and datasets. Here, we report results with and without using the gain multiplier map. This gain multiplier map can be generated by our network by adding an additional decoder network with skip connections from the query encoder. 
Based on this modification, our convolutional structure can now be described as:
\begin{equation}
P = \operatorname{softmax}\bigg(B + G \circ \sum_i\big(N_i * F_i\big)\bigg)\,, \label{eq:CCC_w_G}
\end{equation}
where $\{ F_i \}$, $B$, and $G$ are filters, a bias map $B(i,j)$, and the gain multiplier map $G(i,j)$, respectively. We also change the smoothness regularizer to include the generated gain multiplier as follows:
\begin{align}
S\left(\{ F_i \}, B, G\right) = \lambda_{B} (&\lVert B \ast \sobel_u \rVert^2 + \lVert B \ast \sobel_v \rVert^2 )  \nonumber \\
+ \lambda_{G} (&\lVert G \ast \sobel_u \rVert^2 + \lVert G \ast \sobel_v \rVert^2 ) \nonumber \\
+ \lambda_{F} \sum_i (&\lVert F_i \ast \sobel_u \rVert^2 + \lVert F_i \ast \sobel_v \rVert^2 ) \,, \label{loss:Eq.2}
\end{align}
\noindent where $\sobel_u$ and $\sobel_v$ are $3\!\times\!3$ horizontal and vertical Sobel filters, respectively, and $\lambda_{F}$, $\lambda_{B}$, $\lambda_{G}$ are scalar multipliers to control the strength of the smoothness of each of the filters, the bias, and the gain, respectively. The results of using the additional gain multiplier map are reported in Table \ref{table:with_gain}.

\begin{table*}[t]
\centering
\caption{Results of using the gain multiplier, $G$. For each experiment, we used $m=7$ and $n=64$, and trained our network using the same training data explained in the main paper with augmentation. Lowest errors are highlighted in yellow.\label{table:with_gain}}
\scalebox{0.58}{
\begin{tabular}{l|cccc|cccc|cccc|cccc|cccc}
 & \multicolumn{4}{c|}{\textbf{Cube+} \cite{banic2017unsupervised}} & \multicolumn{4}{c|}{\textbf{Cube+ Challenge} \cite{challenge}} & \multicolumn{4}{c|}{\textbf{INTEL-TAU} \cite{laakom2019intel}} & \multicolumn{4}{c|}{\textbf{Gehler-Shi} \cite{gehler2008bayesian}} & \multicolumn{4}{c}{\textbf{NUS} \cite{cheng2014illuminant}} \\ \cline{2-21} 
 & \textbf{Mean} & \textbf{Med.} & \textbf{B. 25\%} & \textbf{W. 25\%} & \textbf{Mean} & \textbf{Med.} & \textbf{B. 25\%} & \textbf{W. 25\%} & \textbf{Mean} & \textbf{Med.} & \textbf{B. 25\%} & \textbf{W. 25\%} & \textbf{Mean} & \textbf{Med.} & \textbf{B. 25\%} & \textbf{W. 25\%} & \textbf{Mean} & \textbf{Med.} & \textbf{B. 25\%} & \textbf{W. 25\%} \\ \hline
w/o $G$ &  1.87 & 1.27 & \cellcolor[HTML]{\bestcolor} 0.41 & 4.36 &  2.40 & 1.58 & 0.52 & \cellcolor[HTML]{\bestcolor} 5.76 &  \cellcolor[HTML]{\bestcolor}2.49 & \cellcolor[HTML]{\bestcolor} 1.66 & \cellcolor[HTML]{\bestcolor} 0.51 & \cellcolor[HTML]{\bestcolor} 5.93 & \cellcolor[HTML]{\bestcolor} 2.36 & \cellcolor[HTML]{\bestcolor} 1.61 & \cellcolor[HTML]{\bestcolor}0.44 & 5.60 & 2.68 & 2.00 & 0.66 & 5.90 \\
w/ $G$ & \cellcolor[HTML]{\bestcolor}1.83 & \cellcolor[HTML]{\bestcolor}1.24 & 0.42 & \cellcolor[HTML]{\bestcolor}4.25 & \cellcolor[HTML]{\bestcolor}2.34 & \cellcolor[HTML]{\bestcolor}1.45 & \cellcolor[HTML]{\bestcolor}0.46 & 5.86 & 2.63 & 1.81 & 0.55 & 6.18 & \cellcolor[HTML]{\bestcolor}2.36 & 1.72 & 0.48 & \cellcolor[HTML]{\bestcolor}5.40 & \cellcolor[HTML]{\bestcolor}2.44 & \cellcolor[HTML]{\bestcolor}1.89 & \cellcolor[HTML]{\bestcolor}0.64 & \cellcolor[HTML]{\bestcolor}5.21 \\
\end{tabular}}
\end{table*}

We further trained and tested our C5 model using the INTEL-TAU dataset evaluation protocols \cite{laakom2019intel}. Specifically, the INTEL-TAU dataset introduced two different evaluation protocols: (i) the cross-validation protocol, where the model is trained using a 10-fold cross-validation scheme of images taken from three different camera models, and (ii) the camera invariance evaluation protocol, where the model is trained on a single camera model and then tested on another camera model. This camera invariance protocol is equivalent to the CS evaluation method~\cite{afifi2019sensor}, as the models are trained and tested on the same scene set, but with different camera models in the training and testing phases. See Table \ref{table:intel_tau} for comparison with other methods using the INTEL-TAU evaluation protocols. In Table \ref{table:intel_tau}, we also show the results of our C5 model trained on the NUS and Gehler-Shi datasets with augmentation (i.e., our camera-independent model) as reported in Table \ref{table:results} for completeness.


\begin{table}[]
\centering
\caption{Results using the INTEL-TAU dataset evaluation protocols \cite{laakom2019intel}. We also show the results of camera-independent methods, including our camera-independent C5 model. Lower errors for each evaluation protocol are highlighted in yellow. The best results are bold-faced.\label{table:intel_tau}}
\vspace{1mm}
\scalebox{0.65}{
\begin{tabular}{l|ccccc}
\textbf{\begin{tabular}[c]{@{}l@{}}\textbf{INTEL-TAU} \cite{laakom2019intel} \end{tabular}} & \textbf{Mean} & \textbf{Med.} & \textbf{B. 25\%} & \textbf{W. 25\%} & \textbf{Tri.}  \\ \hline
\multicolumn{6}{c}{\cellcolor[HTML]{\headercolor}\textbf{Camera-specific (10-fold cross-validation protocol \cite{laakom2019intel})}} \\ \hline
Bianco et al.'s CNN \cite{bianco2015color}  &  3.5 &  2.6 & 0.9 & 7.4 & 2.8 \\
C3AE \cite{laakom2019color}  &  3.4 & 2.7 & 0.9 & 7.0 &  2.8 \\
BoCF \cite{laakom2020bag}  & 2.4 &  1.9 & 0.7 & 5.1 & 2.0 \\
FFCC \cite{FFCC} &  2.4 &  1.6  & 0.4 & 5.6 & 1.8 \\
VGG-FC$^4$ \cite{hu2017fc}  &  \cellcolor[HTML]{\bestcolor}\textbf{2.2} & 1.7 & 0.6 &  \cellcolor[HTML]{\bestcolor}\textbf{4.7} & 1.8 \\
\hdashline
C5 ($m=7, n=128$), w/ augmentation & 2.33 & \cellcolor[HTML]{\bestcolor}\textbf{1.55} & \cellcolor[HTML]{\bestcolor}\textbf{0.45} & 5.57 & \cellcolor[HTML]{\bestcolor}\textbf{1.71} \\\hline

\multicolumn{6}{c}{\cellcolor[HTML]{\headercolor}\textbf{Camera-specific (camera invariant protocol \cite{laakom2019intel})}} \\ \hline
Bianco et al.'s CNN \cite{bianco2015color} &  3.4 &  2.5 & 0.8 & 7.2 & 2.7 \\

C3AE \cite{laakom2019color} & 3.4 & 2.7 & 0.9 &  7.0 &  2.8 \\

BoCF \cite{laakom2020bag} &  2.9 & 2.4 & 0.9 &  6.1 & 2.5 \\

VGG-FC$^4$ \cite{hu2017fc}  &  2.6 &  2.0 & 0.7 & 5.5 & 2.2 \\

\hdashline

C5 ($m=9$), \waugmentation & \cellcolor[HTML]{\bestcolor}2.45 & \cellcolor[HTML]{\bestcolor}1.82 & \cellcolor[HTML]{\bestcolor}0.53  & \cellcolor[HTML]{\bestcolor}5.46 & \cellcolor[HTML]{\bestcolor}1.95\\

\hline
\multicolumn{6}{c}{\cellcolor[HTML]{\headercolor}\textbf{Camera-independent}} \\ \hline

Gray-world \cite{GW} &  4.7 &  3.7 & 0.9 & 10.0 &  4.0\\
White-Patch \cite{maxRGB} &  7.0  & 5.4 & 1.1 & 14.6 & 6.2\\
1st-order Gray-Edge \cite{maxRGB} & 5.3 & 4.1 & 1.0 & 11.7 & 4.5\\
2nd-order Gray-Edge \cite{maxRGB} & 5.1 & 3.8 & 1.0 & 11.3 & 4.2\\
Shades-of-Gray \cite{SoG} & 4.0 &  2.9 & 0.7 & 9.0 &  3.2\\
PCA-based B/W Colors \cite{cheng2014illuminant} & 4.6 & 3.4 & 0.7 & 10.3 & 3.7 \\
Weighted Gray-Edge \cite{WGE} & 6.0 & 4.2 & 0.9 & 14.2 & 4.8\\
Quasi-Unsupervised CC \cite{bianco2019quasi} & 3.71 & 2.67 & 0.66 & 8.55 & 2.90 \\
SIIE \cite{afifi2019sensor} & 3.42 & 2.42 & 0.73 &  7.80 & 2.64 
\\ 

\hdashline
C5 ($m=7$), \waugmentation & \cellcolor[HTML]{\bestcolor}2.49 & \cellcolor[HTML]{\bestcolor}1.66 & \cellcolor[HTML]{\bestcolor}0.51 & \cellcolor[HTML]{\bestcolor}5.93 & \cellcolor[HTML]{\bestcolor}1.83 \\

\end{tabular}}
\end{table}

Our C5 model achieves reasonable accuracy when used as a camera-specific model. In this scenario, we trained our model on training images captured by the same test camera model with a single encoder (i.e., $m=1$). We found that $n=128$, using the gain multiplier map $G(i,j)$, achieves the best camera-specific results. We report the results of our camera-specific models in Table \ref{table:camera_specific}.



\begin{table}[t]
\centering
\caption{Results of our C5 trained as a \textbf{camera-specific model} with a single encoder (i.e., $m=1$). In this experiment, we performed a three-fold cross-validation on the Cube+ dataset \cite{banic2017unsupervised}. For the Cube+ challenge \cite{challenge}, we report our results after training our model on the Cube+ dataset \cite{banic2017unsupervised} without including any training example from the Cube+ challenge test set \cite{challenge}. We also show the results of other camera-specific color constancy methods reported in past papers. Lowest angular errors are highlighted in yellow. \label{table:camera_specific}}
\vspace{1mm}
\resizebox{\linewidth}{!}{
\begin{tabular}{l|ccccc}
\textbf{\begin{tabular}[c]{@{}l@{}}\textbf{Cube+ Dataset} \cite{banic2017unsupervised} \end{tabular}} &  \textbf{Mean} & \textbf{Med.} & \textbf{B. 25\%} & \textbf{W. 25\%} & \textbf{Tri.}  \\ \hline

Color Dog \cite{banic2015color}& 3.32 & 1.19 & 0.22 & 10.22 & -  \\
APAP \cite{Afifi19}& 2.01  &  1.36 & 0.38 & 4.71 &  -  \\
Meta-AWB w/ 20 tuning images \cite{mcdonagh2018formulating} & 1.59 & 1.02 & 0.30 &  3.85 & 1.15 - \\
Color Beaver \cite{kovsvcevic2019color}& 1.49 & 0.77 & 0.21 &
3.94 & -  \\
SqueezeNet-FC$^4$ \cite{hu2017fc} & 1.35 & 0.93 & 0.30 & 3.24 & 1.01  \\
FFCC \cite{FFCC} & 1.38 & \cellcolor[HTML]{\bestcolor}0.74 & 0.19 & 3.67 & \cellcolor[HTML]{\bestcolor}0.89  \\
WB-sRGB (modified for raw-RGB) \cite{afifi2019color} & 1.32 & \cellcolor[HTML]{\bestcolor}0.74 & \cellcolor[HTML]{\bestcolor}0.18 & 3.43 & -  \\
MDLCC \cite{xiao2020multi} & \cellcolor[HTML]{\bestcolor}1.24 & 0.83 &  0.26 & \cellcolor[HTML]{\bestcolor}2.91 &  0.92 \\
\hdashline
C5 ($n=128$), w/ $G$ & 1.39 & 0.79 & 0.24 & 3.55 & 0.93  \\
\end{tabular}}

\vspace{0.16in}

\resizebox{\linewidth}{!}{
\begin{tabular}{l|ccccc}
\textbf{\begin{tabular}[c]{@{}l@{}}\textbf{Cube+ Challenge} \cite{challenge}\end{tabular}} & \textbf{Mean} & \textbf{Med.} & \textbf{B. 25\%} & \textbf{W. 25\%} & \textbf{Tri.}  \\ \hline
V Vuk et al., \cite{challenge}& 6.00 & 1.96 & 0.99 & 18.81 & 2.25 \\
A Savchik et al., \cite{savchik2019color} & 2.05 & 1.20 & 0.40 & 5.24 & 1.30 \\
Y Qian et al., (1) \cite{qian2019fast} &  2.48 & 1.56 & 0.44 & 6.11 & - \\
Y Qian et al., (2) \cite{qian2019fast} & 2.27 & 1.26  & 0.39 & 6.02 & 1.35\\
FFCC \cite{FFCC} &  2.1 & 1.23 & 0.47 & 5.38 & - \\
MHCC \cite{hernandez2020multi} & 1.95 & 1.16 & 0.39 & 4.99 & 1.25 \\
K Chen et al., \cite{challenge}& 1.84 & 1.27 & 0.39 & 4.41 & 1.32 \\
WB-sRGB (modified for raw-RGB) \cite{afifi2019color}  & 1.83 & 1.15 & \cellcolor[HTML]{\bestcolor}0.35 & 4.60 & - \\
\hdashline
C5 ($n=128$), w/ $G$ & \cellcolor[HTML]{\bestcolor}1.72 & \cellcolor[HTML]{\bestcolor}1.07 & 0.36 & \cellcolor[HTML]{\bestcolor}4.27 & \cellcolor[HTML]{\bestcolor}1.15\\
\end{tabular}}

\end{table}

\begin{figure*}[t]
\centering
\includegraphics[width=\linewidth]{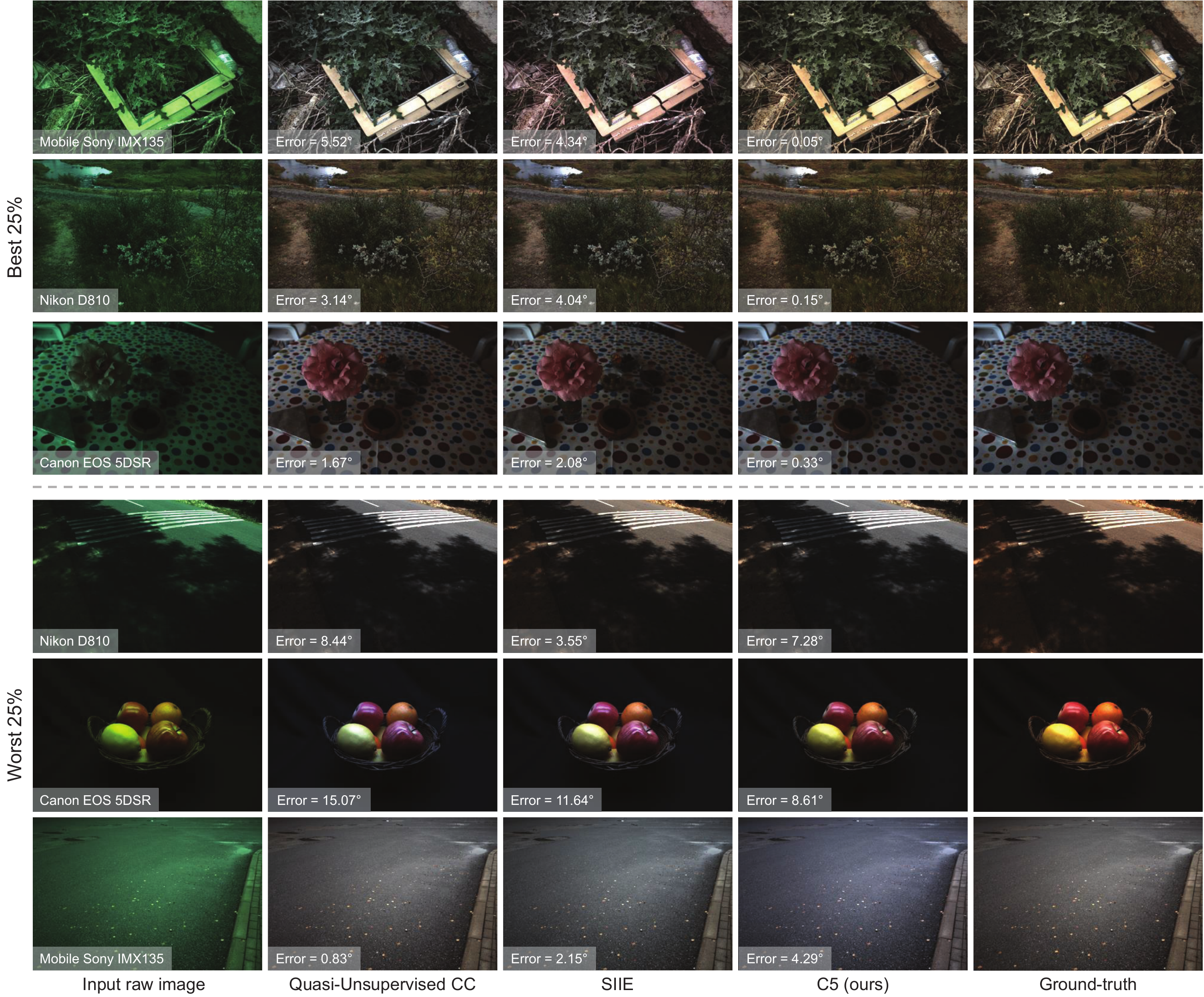}
\vspace{-12pt}
\caption{\small{Random examples from our ``worst 25\%'' and ``best 25\%'' results alongside quasi-unsupervised CC \cite{bianco2019quasi} and SIIE \cite{afifi2019sensor}. Input images are from the INTEL-TAU dataset \cite{laakom2019intel}. 
\label{fig:qualitative2}}}
\vspace{-15pt}
\end{figure*}

Lastly, we show additional qualitative results from the INTEL-TAU dataset \cite{laakom2019intel} in Figure \ref{fig:qualitative2}. In this figure, we show qualitative examples from our ``worst 25\%'' and ``best 25\%'' results alongside the corresponding results of prior sensor-independent
techniques \cite{afifi2019sensor, bianco2019quasi}.

\section{Data Augmentation} \label{sec:augmentation}
In this section, we describe in detail the data augmentation procedure described in Sec.\ \ref{subsec:dataaug}. We begin with the steps used to map a color temperature to the corresponding CIE XYZ value. Then, we elaborate the process of mapping from camera sensor raw to the CIE XYZ color space. Afterwards, we describe the details of the scene retrieval process mentioned in Sec.\ \ref{subsec:dataaug}. Finally, we discuss experiments performed to evaluate our data augmentation and compare it with other color constancy augmentation techniques used in the literature.

\subsection{From Color Temperature to CIE XYZ} \label{subsec:temp_to_xyz}

According to Planck's radiation law \cite{wyszecki1982color}, the spectral power distribution (SPD) of a blackbody radiator at a given wavelength range $[\lambda, \partial\lambda]$ can be computed using the color temperature $q$ as follows:

\begin{equation}\label{xyz_conv:Eq.1}
S_{\lambda}d_{\lambda}= \frac{f_1\lambda^{-5}}{\exp{(f_2/\lambda q}) - 1}\partial\lambda,
\end{equation}

\noindent  where, $f_1= 3.741832^10^{-16}$ $Wm^2$ is the first radiation constant,  $f_2=1.4388^{10-2} mK$ is the second radiation constant, and $q$ is the blackbody temperature, in Kelvin. \cite{johnwalker, li2016accurate}. Once the SPD is computed, the corresponding CIE tristimulus values can be approximated in the following discretized form:

\begin{equation} \label{xyz_conv:Eq.2}
X =\Delta\lambda \sum_{\lambda=380}^{\lambda=780}x_{\lambda} S_{\lambda},
\end{equation}

\noindent where the value of $x_{\lambda}$ is the standard CIE color match value \cite{cie1932commission}. The values of $Y$ and $Z$ are computed similarly. The corresponding chromaticity coordinates of the computed XYZ tristimulus are finally computed as follows:
\begin{equation}
\begin{aligned} \label{xyz_conv:Eq.3}
x = X/(X+Y+Z),\\
y = Y/(X+Y+Z),\\
z = Z/(X+Y+Z).\\
\end{aligned}
\end{equation}

\subsection{From Raw to CIE XYZ} \label{subsec:raw_to_xyz}

Most DSLR cameras provide two pre-calibrated matrices, $C_1$ and $C_2$, to map from the camera sensor space to the CIE 1931 XYZ 2-degree standard observer color space. These pre-calibrated color space transformation (CST) matrices are usually provided as a low color temperature (e.g., Standard-A) and a higher correlated color temperature (e.g., D65) \cite{DNG}.  

Given an illuminant vector $\mat{\light}$, estimated by an illuminant estimation algorithm, the CIE XYZ mapping matrix associated with $\mat{\light}$ is computed as follows \cite{can2018improving}:

\begin{equation}\label{raw_to_xyz_conv:Eq.1}
C_{T_{\mat{\light}}}=\alpha C_1 + \left(1-\alpha\right)C_2,
\end{equation}
\begin{equation}\label{raw_to_xyz_conv:Eq.2}
\alpha=(1/q_{\mat{\light}} - 1/q_2)/(1/q_1 - 1/q_2),
\end{equation}

\noindent  where $q_1$ and $q_2$ are the correlated color temperature associated to the pre-calibrated matrices $C_1$ and $C_2$, and $q_{\mat{\light}}$ is the color temperature of the illuminant vector $\mat{\light}$. Here, $q_{\mat{\light}}$ is unknown, and unlike the standard mapping from color temperature to the CIE XYZ space (Sec.~\ref{subsec:temp_to_xyz}), there is no standard conversion from a camera sensor raw space to the corresponding color temperature. Thus, the conversion from the sensor raw space to the CIE XYZ space is a chicken-and-egg problem---computing the correlated color temperature $q_{\mat{\light}}$ is necessarily to get the CST matrix $C_{q_{\mat{\light}}}$, while knowing the mapping from a camera sensor raw to the CIE XYZ space inherently requires knowledge of the correlated color temperature of a given raw illuminant.

This problem can be solved by a trial-and-error strategy as follows. We iterate over the color temperature range of 2500K to 7500K. For each color temperature $q_i$ , we first compute the corresponding CST matrix $C_{q_i}$ using Eqs. \ref{raw_to_xyz_conv:Eq.1} and \ref{raw_to_xyz_conv:Eq.2}. Then, we convert $q_i$ to the corresponding xyz chromaticity triplet using Eqs. \ref{xyz_conv:Eq.1}--\ref{xyz_conv:Eq.3}. 

Afterwards, we map the xyz chromaticity triplet to the sensor raw space using the following equation:

\begin{equation}\label{raw_to_xyz_conv:Eq.3}
\mat{\light}_{\texttt{raw}(q_i)} = C_{q_i}^{-1} \lambda_{\texttt{xyz}(q_i)}.
\end{equation}

We repeated this process for all color temperatures and selected the color temperature/CST matrix that achieves the minimum angular error between $\mat{\light}$ and the reconstructed illuminant color in the sensor raw space. 

The accuracy of our conversion depends on the pre-calibrated matrices provided by the manufacturer of the DSLR cameras. Other factors that may affect the accuracy of the mapping includes the precision of the standard mapping from color temperature to XYZ space defined by \cite{cie1932commission}, and the discretization process in Eq. \ref{xyz_conv:Eq.2}.

\subsection{Raw-to-raw mapping}

Here, we describe the details of the mapping mentioned in Sec.\ \ref{subsec:dataaug}. Let $A$=$\{\mat{a}_1, \mat{a}_2, ...\}$ represent the ``source'' set of demosaiced raw images taken by different camera models with the associated capture metadata. Let $T=\{\mat{t}_1, \mat{t}_2 , ...\}$ represent our ``target'' set of metadata of captured scenes by the target camera model. Here, the capture metadata includes exposure time, aperture size, ISO gain value, and the global scene illuminant color in the camera sensor space. We also assume that we have access to the pre-calibration color space transformation (CST) matrices for each camera model in the sets $A$ and $T$ (available in most DNG files of DSLR images \cite{DNG}).

Our goal here is to map all raw images in $A$, taken by different camera models, to the target camera sensor space in $T$. To that end, we map each image in $A$ to the device-independent CIE XYZ color space \cite{cie1932commission}. This mapping is performed as follows. We first compute the correlated color temperature, $q^{(i)}$, of the scene illuminant color vector, $\mat{\light}^{(i)}_{\texttt{raw}(A)}$, of each raw image, $I^{(i)}_{\texttt{raw}(A)}$, in the set $A$ (see Sec. \ref{subsec:raw_to_xyz}). 
Then, we linearly interpolate between the pre-calibrated CST matrices provided with each raw image to compute the final CST mapping matrix, $C_{q^{(i)}}$, \cite{can2018improving}. 
Afterwards, we map each image, $I^{(i)}_{\texttt{raw}(A)}$, in the set $A$ to the CIE XYZ space. Note that here we represent each image $I$ as matrices of the color triplets (i.e., $I = \{\mat{c}^{(k)}\}$), where $k$ is the total number of pixels in the image $I$. We map each raw image to the CIE XYZ space as follows:

\begin{equation}\label{data_aug:Eq.1}
I^{(i)}_{\texttt{xyz}(A)}= C_{q^{(i)}}D_{\mat{\light}^{(i)}}I^{(i)}_{\texttt{raw}(A)},
\end{equation}

\noindent where  $D_{\mat{\light}^{(i)}}$ is the white-balance diagonal correction matrix constructed based on the illuminant vector $\mat{\light}^{(i)}_{\texttt{raw}(A)}$.

Similarly, we compute the inverse mapping from the CIE XYZ space back to the target camera sensor space based on the illuminant vectors and pre-calibration matrices provided in the target set $T$. The mapping from the source sensor space to the target one in $T$ can be performed as follows:

\begin{equation}\label{data_aug:Eq.2}
I^{(i)}_{\texttt{raw}(T)}= D^{-1}_{	\jmath^{(i)}}M^{-1}_{q^{(i)}}I^{(i)}_{\texttt{xyz}(A)},
\end{equation}

\noindent where $\jmath^{(i)}_{\texttt{raw}(T)}$ is the corresponding illuminant color to the correlated color temperature, $q^{(i)}$, in the target sensor space (i.e., the ground-truth illuminant for image $I^{(i)}_{\texttt{raw}(T)}$ in the illuminant estimation task), and $M^{-1}_{q^{(i)}}$ is the CST matrix that maps from the target sensor space to the CIE XYZ space. 

The described steps so far assume that the spectral sensitivities of all sensors in $A$ and $T$ satisfy the Luther condition \cite{nakamura2017image}. Prior studies, however, showed that this assumption is not always satisfied, and this can affect the accuracy of the pre-calibration matrices \cite{jiang2013space, karaimer2019beyond}. According to this, we rely on Eqs. \ref{data_aug:Eq.1} and \ref{data_aug:Eq.2} only to map the original colors of captured objects in the scene (i.e., white-balanced colors) to the target camera model. For the values of the global color cast, $\jmath^{(i)}_{\texttt{raw}(T)}$, we do not rely on $M^{-1}_{q^{(i)}}$ to map $\mat{\light}^{(i)}_{\texttt{raw}(A)}$ to the target sensor space of $T$. Instead, we follow a $K$-nearest neighbor strategy to get samples from the target sensor's illuminant color space.

\subsection{Scene Sampling} \label{subsec:scene_retrieval}
As described in Sec.\ \ref{subsec:dataaug}, we retrieve metadata of similar scenes in the target set $T$ for illuminant color sampling. This sampling process should consider the source scene capture conditions to sample suitable illuminant colors from the target camera model space---i.e., having indoor illuminant colors as ground-truth for outdoor scenes may affect the training process. To this end, we introduce a retrieval feature $v_A^{(i)}$ to represent the capture settings of the image $I^{(i)}_{\texttt{raw}(A)}$. This feature includes the correlated color temperature and auxiliary capture settings. These additional capture settings are used to retrieve scenes captured with similar settings of $I^{(i)}_{\texttt{raw}(A)}$. 

Our feature vector is defined as follows:
\begin{equation}\label{data_aug:Eq.3}
v_A^{(i)} = [q_{\texttt{norm}}^{(i)} \,,  h_{\texttt{norm}}^{(i)}, p_{\texttt{norm}}^{(i)}\,, e_{\texttt{norm}}^{(i)}],  
\end{equation}
where $q_{\texttt{norm}}^{(i)}$, $ h_{\texttt{norm}}^{(i)}$, $p_{\texttt{norm}}^{(i)}$, and $e_{\texttt{norm}}^{(i)}$ are the normalized color temperature, gain value, aperture size, and scaled exposure time, respectively. The gain value and the scaled exposure time are computed as follows:
\begin{equation}\label{data_aug:Eq.4}
h^{(i)} = \texttt{BLN}^{(i)} \texttt{ISO}^{(i)}\,,  
\end{equation}
\begin{equation}\label{data_aug:Eq.5}
e^{(i)} = \sqrt{2^{\texttt{BLE}^{(i)}}} l^{(i)}\,, 
\end{equation}
where $\texttt{BLE}$, $\texttt{BLN}$, $\texttt{ISO}$, and $l$ are the baseline exposure, baseline noise, digital gain value, and exposure time (in seconds), respectively.

\paragraph{Illuminant Color Sampling}

A naive sampling from the associated illuminant colors in $T$ does not introduce new illuminant colors over the Planckian locus of the target sensor. For this reason, we first fit a cubic polynomial to the $rg$ chromaticity of illuminant colors in the target sensor $T$. Then, we compute a new $r$ chromaticity value for each query vector as follows:
\begin{equation}\label{data_aug:Eq.6}
\centering
r_v = \sum_{j\in K}{w_jr_j + x}\,,
\end{equation}
where $w_j = \exp(1-d_j)/\sum_{k}^{K}{\exp(1-d_k)}$ is a weighting factor,  $x=\lambda_r \mathcal{N}(0, \sigma_r)$ is a small random shift, $\lambda_r$ is a scalar factor to control the amount of divergence from the ideal Planckian curve, $\sigma_r$ is the standard deviation of the $r$ chromaticity values in the retrieved $K$ metadata of the target camera model, $T_K$, and $d_j$ is the normalized L2 distance between $v_{S(i)}$ and the corresponding $j^\texttt{th}$ feature vector in $T_K$. The CST matrix $M$ (Eq. \ref{data_aug:Eq.2}) is constructed by linearly interpolating between the corresponding CST matrices associated with each sample in $T_K$ using $w_j$. After computing $r_v$, the corresponding $g$ chromaticity value is computed as:
\begin{equation}\label{data_aug:Eq.8}
g_v=[r_v, r^2_v, r^3_v] [\xi_1, \xi_2, \xi_3]^\top + y\,,   
\end{equation}
where $[\xi_1, \xi_2, \xi_3]$ are the cubic polynomial coefficients, $y$ is a random shift, and $\sigma_g$ is the standard deviation of the $g$ chromaticity values in $T_K$. In our experiments, we set $\lambda_r = 0.7$ and $\lambda_g = 1$. The final illuminant color $\jmath^{(i)}_{\texttt{raw}(T)}$ can be represented as follows:
\begin{equation}\label{data_aug:Eq.9}
\jmath^{(i)}_{\texttt{raw}(T)} = [r_v, g_v, 1 - r_v- g_v]^\top\,.
\end{equation}

\begin{figure}
\centering
\includegraphics[width=\linewidth]{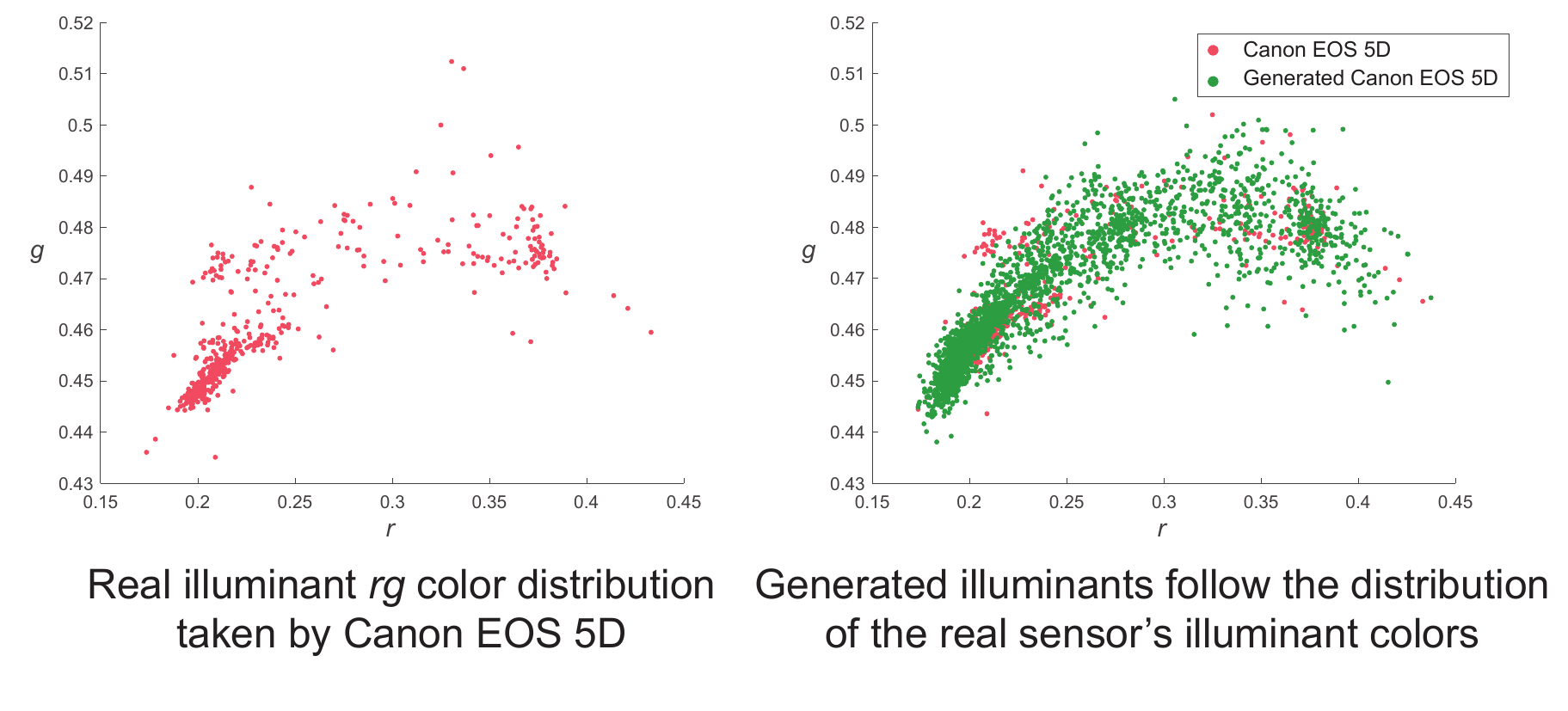}
\vspace{-6mm}
\caption{Synthetic illuminant samples of Canon EOS 5D camera model in the Gehler-Shi dataset \cite{gehler2008bayesian}. The shown generated illuminant colors are then applied to sensor-mapped raw images, originally were taken by different camera models, for augmentation purpose (Sec. \ref{sec:augmentation}).}
\label{fig:synthetic_ill_plot}
\end{figure}

\begin{figure}
\centering
\includegraphics[width=\linewidth]{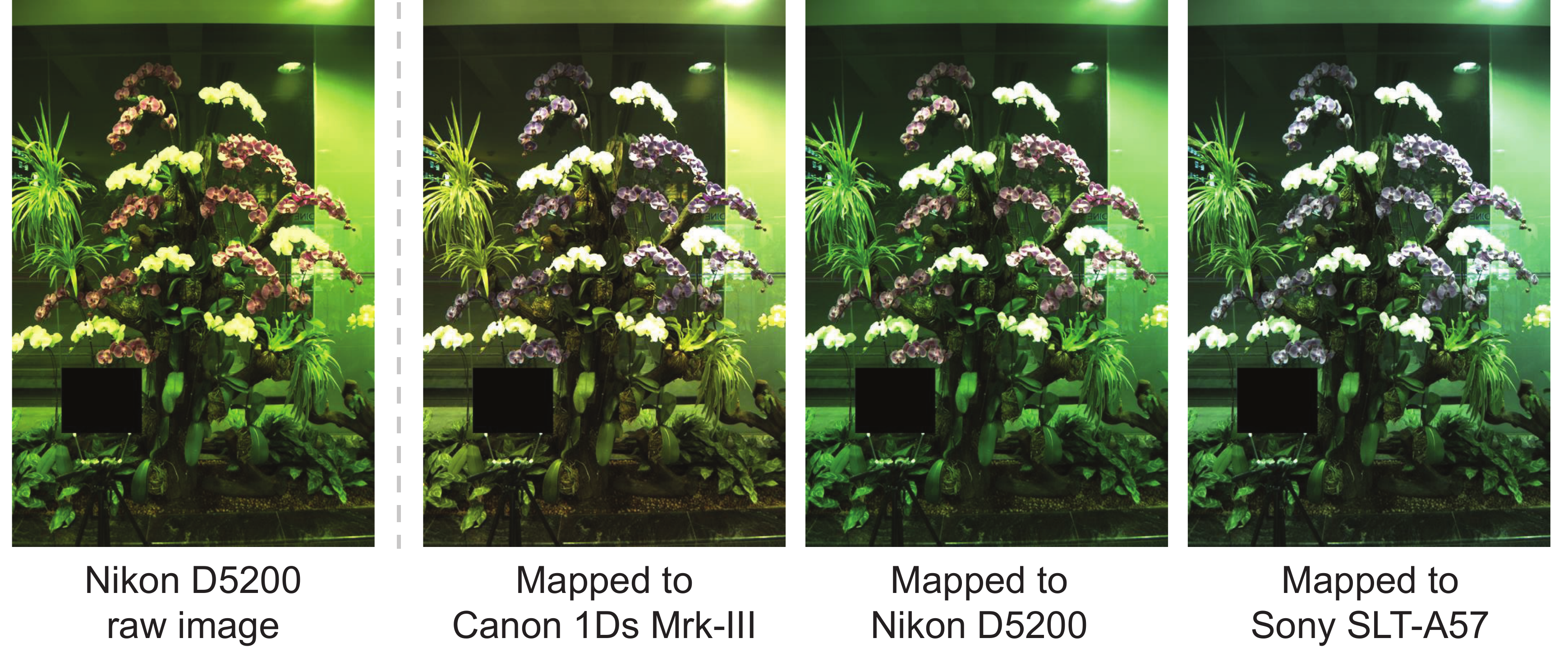}
\vspace{-6mm}
\caption{Example of camera augmentation used to train our network. The shown left raw image is captured by Nikon D5200 camera \cite{cheng2014illuminant}. The next three images are the results of our mapping to different camera models. \label{fig:augmented_images_example}}

\end{figure}

To avoid any bias towards the dominant color temperature in the source set, $A$, we first divide the color temperature range of the source set $A$ into different groups with a step of 250K. Then, we uniformly sample examples from each group to avoid any bias towards specific type of illuminants. Figure \ref{fig:synthetic_ill_plot} shows examples of the sampling process. As shown, the sampled illuminant chromaticity values follow the original distribution over the Planckian curve, while introducing new illuminant colors of the target sensors that were not included in the original set. Finally, we apply random cropping to introduce more diversity in the generated images. Figure \ref{fig:augmented_images_example} shows examples of synthetic raw-like images of different target camera models.

\subsection{Evaluation} \label{subsec:synthetic_data_eval}

In prior work, several approaches for training data augmentation for illuminant estimation have been attempted \cite{bmvc1, fourure2016mixed, afifi2020cie}. These approaches first white-balance the training raw images using the associated ground-truth illuminant colors associated with each image. Afterwards, illuminant colors are sampled from the ``ground-truth'' illuminant colors over the entire training set to be applied to the white-balanced raw images. These sampled illuminant colors can be taken randomly from the ground-truth illuminant colors \cite{fourure2016mixed} or after clustering the ground-truth illuminant colors \cite{bmvc1}. These methods, however, are limited to using the same set of scenes as is present in the training dataset. Another approach for data augmentation has been proposed in \cite{afifi2020cie} by mapping sRGB white-balanced images to a learned normalization space that is is learned based on the CIE XYZ space. Afterwards, a pre-computed global transformation matrix is used to map the images from this normalization space to the target white-balanced raw space. In contrast, the augmentation method described in our paper uses an accurate mapping from the camera sensor raw space to the CIE XYZ using the pre-calibration matrices provided by camera manufacturers. 

In the following set of experiments, we use the baseline model FFCC \cite{FFCC} to study the potential improvement of our chosen data augmentation strategy and alternative augmentation techniques proposed in \cite{bmvc1, fourure2016mixed, afifi2020cie}. We use the Canon EOS 5D images from in the Gehler-Shi dataset \cite{gehler2008bayesian} for comparisons. For our test set, we randomly select 30\% of the total number of images in the Canon EOS 5D set. The remaining 70\% of images are used for training. We refer to this set as ``real training set'', which includes 336 raw images.

Note that, except for the augmentation used in a  \cite{afifi2020cie}, none of these methods apply a sensor-to-sensor mapping, as they use the raw images of the ``real training set'' as the source and target set for augmentation. For this reason and for a fair comparison, we provide the results of two different set of experiments. In the first experiment, we use the CIE XYZ images taken by the Canon EOS 5D sensor as our source set $A$, while in the second experiment, we use a different set of four sensors rather than the Canon EOS 5D sensor. The former is comparable to the augmentation methods used in \cite{bmvc1, fourure2016mixed} (see Table \ref{table:augmentation_1}), while the latter is comparable to the augmentation approach used in \cite{afifi2020cie}, which performs ``raw mapping'' in order to introduce new scene content in the training data (see Table \ref{table:augmentation_2}). The shown results obtained by generating 500 synthetic images by each augmentation method, including our augmentation approach. As shown in Tables \ref{table:augmentation_1} and \ref{table:augmentation_2}, our augmentation approach achieves the best improvement of the FFCC results.

In order to study the effect of the CIE XYZ mapping used by our augmentation approach, we trained FFCC \cite{FFCC} on a set of 500 synthetic raw images of the target camera model---namely, the Canon EOS 5D camera model in the Gehler-Shi dataset \cite{gehler2008bayesian}. These synthetic raw images were originally captured by the Canon EOS 1Ds Mark III camera sensor (in the NUS dataset \cite{cheng2014illuminant}), then these images are mapped to the target sensor using our augmentation approach. Table \ref{table:CIEXYZ_ablation} shows the results of FFCC trained on synthetic raw images with and without the intermediate CIE XYZ mapping step (Eqs. \ref{data_aug:Eq.1} and \ref{data_aug:Eq.2}). As shown, using the CIE XYZ mapping achieves better results, which are further improved by increasing the scene diversity of the source set by including additional scenes from other datasets, as shown in Table \ref{table:augmentation_2}.

For a further evaluation, we use our approach to map images from the Canon EOS 5D camera's set (the same set that was used to train the FFCC model) to different target camera models. Then, we trained and tested a FFCC model on these mapped images. This experiment was performed to gauge the ability of our data augmentation approach to have similar negative effects on camera-specific methods that were trained on a different camera model. To that end, we randomly selected 150 images from the Canon EOS 5D sensor set, which was used to train the FFCC model, as our source image set $A$. Then, we mapped these images to different target camera models using our approach. That means that the training and our synthetic testing set share the same scene content. We report the results in Table \ref{table:sensor_errors}. We also report the testing results on real image sets captured by the same target camera models. As shown in Table \ref{table:sensor_errors}, both real and synthetic sets negatively affect the accuracy of the FFCC model (see Table \ref{table:augmentation_1} for results of the FFCC on a testing set taken by the same training sensor).

\begin{table}[]
\caption{A comparison of different augmentation methods for illuminant estimation. All results were obtained by using training images captured by the Canon EOS 5D camera model \cite{gehler2008bayesian} as the source and target sets for augmentation. Lowest errors are highlighted in yellow.\label{table:augmentation_1}}
\vspace{-1mm}
\centering
\scalebox{0.68}{
\begin{tabular}{l|c|c|c|c}
\textbf{Training set} & \textbf{Mean} & \textbf{Med.} & \textbf{B. 25\%}& \textbf{W. 25\%}\\ \hline
Original set  & 1.81 & 1.12 & 0.35 & 4.43 \\ 
Augmented (clustering \& sampling) \cite{bmvc1}  & 1.68 & \cellcolor[HTML]{\bestcolor}0.97 & \cellcolor[HTML]{\bestcolor}0.25	& 4.31\\
Augmented (sampling) \cite{fourure2016mixed}  & 1.79 & 1.09 & 0.33 & 4.34 
 \\ \hdashline
Augmented (ours)  & \cellcolor[HTML]{\bestcolor}1.55 & 0.98 & 0.28 & \cellcolor[HTML]{\bestcolor}3.68 \\

\end{tabular}}
\end{table}

\begin{table}[]
\caption{A comparison of techniques for generating new sensor-mapped raw-like images that were originally captured by different sensors than the training camera model. The term `synthetic' refers to training FFCC \cite{FFCC} without including any of the original training examples, while the term `augmented' refers to training on synthetic and real images. The best results are bold-faced. Lowest errors of synthesized and augmented sets are highlighted in red and yellow, respectively.\label{table:augmentation_2}}
\vspace{-1mm}
\centering
\scalebox{0.7}{
\begin{tabular}{l|c|c|c|c}

\textbf{Training set} & \textbf{Mean} & \textbf{Med.} & \textbf{B. 25\%} & \textbf{W. 25\%} \\ \hline
Synthetic \cite{afifi2020cie}  & 4.17 & 3.06 & 0.78 & 9.39\\ 
Augmentation \cite{afifi2020cie} & 2.64 & 1.95 & 0.45 & 5.97\\ \hdashline
Synthetic (ours)  & \cellcolor[HTML]{\secondbestcolor} 2.44 & \cellcolor[HTML]{\secondbestcolor} 1.89 & \cellcolor[HTML]{\secondbestcolor} 0.42 & \cellcolor[HTML]{\secondbestcolor} 5.40\\  
Augmented (ours)  & \cellcolor[HTML]{\bestcolor} \textbf{1.75} & \cellcolor[HTML]{\bestcolor} \textbf{1.28} & \cellcolor[HTML]{\bestcolor} \textbf{0.35} & \cellcolor[HTML]{\bestcolor} \textbf{4.15} \\ 
\end{tabular}}
\end{table}

\begin{table}[]

\caption{Results of FFCC \cite{FFCC} trained on synthetic raw-like images after they are mapped to the target camera model. In this experiment, the raw images are mapped from the Canon EOS-1Ds Mark III camera sensor (taken from the NUS dataset \cite{cheng2014illuminant}) to the target Canon EOS 5D camera in the Gehler-Shi dataset \cite{gehler2008bayesian}. The shown results were obtained with and without the intermediate CIE XYZ mapping step to generate the synthetic training set. Lowest errors are highlighted in yellow.\label{table:CIEXYZ_ablation}}
\vspace{-1mm}
\centering
\scalebox{0.72}{
\begin{tabular}{l|c|c|c|c}

\textbf{Synthetic training set} & \textbf{Mean} & \textbf{Med.} & \textbf{B. 25\%}& \textbf{W. 25\%} \\ \hline
w/o CIE XYZ  & 3.30 & 2.55 & 0.60 & 7.21 \\ 
w/ CIE XYZ & \cellcolor[HTML]{\bestcolor}3.04 &\cellcolor[HTML]{\bestcolor}2.36 & \cellcolor[HTML]{\bestcolor}0.56 & \cellcolor[HTML]{\bestcolor}6.58 \\ 
\end{tabular}}
\end{table}

\begin{table}[]
\caption{Results of FFCC \cite{FFCC} trained on the Canon EOS 5D camera  \cite{gehler2008bayesian} and tested on images taken by different camera models from the NUS dataset \cite{cheng2014illuminant} and the Cube+ challenge set \cite{banic2017unsupervised}. The synthetic sets refer to testing images generated by our data augmentation approach, where these images were mapped from the Canon EOS 5D set (used for training) to the target camera models. \label{table:sensor_errors}}
\vspace{-1mm}
\centering
\scalebox{0.7}{
\begin{tabular}{l|c|c|c|c|c|c}

\multirow{2}{*}{\textbf{Testing sensor}} & \multicolumn{3}{c|}{\textbf{Real camera images}} & \multicolumn{3}{c}{\textbf{Synthetic camera images}} \\ \cline{2-7} 
 & \textbf{Mean} & \textbf{Med.} & \textbf{Max} & \textbf{Mean} & \textbf{Med.} & \textbf{Max} \\ \hline
Canon EOS 1D \cite{gehler2008bayesian}& 3.88 & 2.66 & 16.32 & 4.68 & 3.80 & 22.83 \\ 
Fujifilm XM1 \cite{cheng2014illuminant} & 4.22 & 3.05 & 47.87 & 2.91 & 2.06 & 38.93 \\
Nikon D5200 \cite{cheng2014illuminant}& 4.45 & 3.45 & 36.762 & 3.36  & 2.10 & 41.23 \\
Olympus EPL6 \cite{cheng2014illuminant}& 4.35 & 3.56 & 19.89 & 3.28 & 2.27 & 38.81  \\
Panasonic GX1 \cite{cheng2014illuminant}& 2.83 & 2.03 & 16.58 & 3.24 & 2.29 & 17.07 \\
Samsung NX2000 \cite{cheng2014illuminant}& 4.41 & 3.73 & 17.69 & 3.44 & 2.64 & 18.79 \\
Sony A57 \cite{cheng2014illuminant}& 3.84 & 3.02 & 19.38 & 3.04	 & 1.34 & 39.67 \\
Canon EOS 550D \cite{banic2017unsupervised}& 3.83 & 2.49 & 46.55 & 3.14 & 1.98 & 36.30 \\ 
\end{tabular}}
\end{table}

{\small
\bibliographystyle{ieee_fullname}
\bibliography{ref}
}

\end{document}